\documentclass{article}

\usepackage[final]{neurips_2025}

\usepackage[utf8]{inputenc} % allow utf-8 input
\usepackage[T1]{fontenc}    % use 8-bit T1 fonts
\usepackage{hyperref}       % hyperlinks
\usepackage{url}            % simple URL typesetting
\usepackage{booktabs}       % professional-quality tables
\usepackage{amsfonts}       % blackboard math symbols
\usepackage{nicefrac}       % compact symbols for 1/2, etc.
\usepackage{microtype}      % microtypography
\usepackage{xcolor}         % colors
\usepackage{graphicx}
\usepackage{xspace}
\usepackage{amsmath}
\usepackage{multirow}
\usepackage{array}
\usepackage{tabularx}
\usepackage{colortbl}
\usepackage{svg}
\usepackage{graphicx}
\usepackage{subcaption}
\usepackage{soul}
\usepackage{amssymb}
\usepackage{wrapfig}
\usepackage{algorithm}
\usepackage{algpseudocode}
\usepackage{hyperref} 

\newcommand{\methodname}{SEAgent\xspace}
\newcommand{\judge}{World State Model\xspace}

\makeatletter
\DeclareRobustCommand\onedot{\futurelet\@let@token\@onedot}
\def\@onedot{\ifx\@let@token.\else.\null\fi\xspace}

\def\ie{\emph{i.e}\onedot}

\makeatother

\title{SEAgent: Self-Evolving Computer Use Agent with Autonomous Learning from Experience}

\author{
    Zeyi Sun$^{1,2}$ \quad
    Ziyu Liu$^{1,2}$ \quad
    Yuhang Zang$^{2}$ \quad
    \textbf{Yuhang Cao$^{2}$} \quad \\
    \textbf{Xiaoyi Dong$^{2,3}$} \quad
    \textbf{Tong Wu$^{3}$} \quad
    \textbf{Dahua Lin$^{2,3}$} \quad
    \textbf{Jiaqi Wang$^{2}$} \\
    \and % 使用 \and 比手动空两行更好
    $^{1}$Shanghai Jiao Tong University \quad
    $^{2}$Shanghai Artificial Intelligence Laboratory \\
    $^{3}$The Chinese University of Hong Kong \\
    \texttt{szy2023@sjtu.edu.cn, wangjiaqi@pjlab.org.cn} \\
    \url{https://github.com/SunzeY/SEAgent}
}

\begin{document}

\maketitle
\begin{abstract}
Repurposing large vision-language models (LVLMs) as computer use agents (CUAs) has led to substantial breakthroughs, primarily driven by human-labeled data. However, these models often struggle with novel and specialized software, particularly in scenarios lacking human annotations. To address this challenge, we propose \methodname, an agentic self-evolving framework enabling CUAs to autonomously evolve through interactions with unfamiliar software. Specifically, \methodname empowers computer-use agents to autonomously master novel software environments via experiential learning, where agents explore new software, learn through iterative trial-and-error, and progressively tackle auto-generated tasks organized from simple to complex.
To achieve this goal, we design a World State Model for step-wise trajectory assessment, along with a Curriculum Generator that generates increasingly diverse and challenging tasks. The agent's policy is updated through experiential learning, comprised of adversarial imitation of failure actions and Group Relative Policy Optimization (GRPO) on successful ones.
Furthermore, we introduce a specialist-to-generalist training strategy that integrates individual experiential insights from specialist agents, facilitating the development of a stronger generalist CUA capable of continuous autonomous evolution. This unified agent ultimately achieves performance surpassing ensembles of individual specialist agents on their specialized software. We validate the effectiveness of \methodname across five novel software environments within OS-World. Our approach achieves a significant improvement of 23.2\% in success rate, from 11.3\% to 34.5\%, over a competitive open-source CUA, \ie, UI-TARS. 
%and surpasses the leading autonomous reinforcement learning approaches with generalized advantage estimation (21.8\%).

\end{abstract}

\section{Introduction}

\textit{“A new generation of agents will acquire superhuman capabilities by learning predominantly from experience.”~\cite{silver2025welcome}  }  
\begin{flushright}
\textit{--- David Silver, Richard S. Sutton}
\end{flushright}

% With the rapid development of large vision-language models (LVLMs)~\cite{touvron2023llama,grattafiori2024llama,bai2025qwen2,wang2024qwen2,gpt4,claude-3-7,geminiteam2024geminifamilyhighlycapable,team2023gemini}, computer use agents (CUAs)~\cite{claude,operator,qin2025uitars,lin2024showui,wu2024atlas} have not only emerged but also gained practical utility. Leveraging the powerful perception and reasoning capabilities of LVLMs, these agents can take screenshots as visual input and control computers through keyboard and mouse actions. These CUAs have the potential to simplify human life by acting as versatile assistants and automating human workflows.

With the rapid development of large vision-language models (LVLMs)~\cite{touvron2023llama,grattafiori2024llama,bai2025qwen2,wang2024qwen2,gpt4,claude-3-7,team2023gemini}, computer use agents (CUAs)~\cite{claude,operator,qin2025uitars,lin2024showui,wu2024atlas} have not only emerged but also demonstrated increasing practical utility. By leveraging the powerful perception and reasoning capabilities of LVLMs, these agents can interpret screenshots as visual inputs and operate computers via keyboard and mouse actions. 
%CUAs hold significant potential to enhance daily life by serving as versatile assistants and automating a wide range of human workflows. 
Despite their promising capabilities, current CUAs~\cite{qi2024webrl,putta2024agent,deng2023mind2web,he2024webvoyager,bai2024digirl,lu2024gui} primarily depend on costly human-curated datasets~\cite{deng2023mind2web,chen2024guicourse,wu2024atlas,kapoor2024omniact,li2024effects}, which are typically derived from demonstrations~\cite{lu2024gui,zhang2023you,gur2023real,rawles2023androidinthewild,zhang2024agentohana} or video tutorials in the wild~\cite{xu2024agenttrek}. However, new software continuously emerges and existing software may regularly be updated, often in the absence of annotated human data. It is both necessary and timely to enter an era that emphasizes learning from experience~\cite{silver2025welcome} in CUA domain. In this paper, we aim to enable CUAs to autonomously explore unfamiliar software environments and evolve into experts without relying on human supervision.

\begin{figure*}[t]
  \centering
  \includegraphics[width=1.0\linewidth]{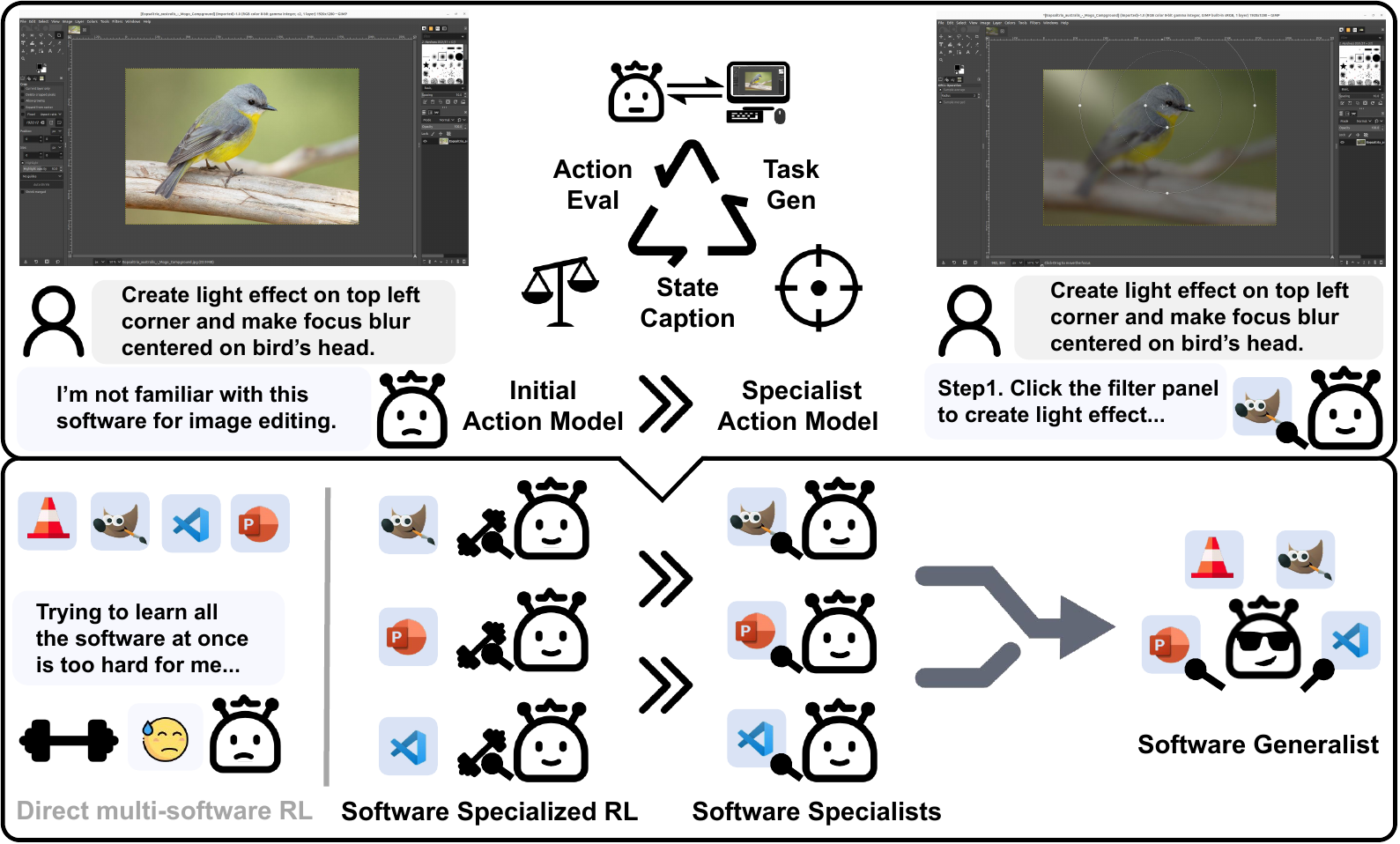}
  \vspace{-2mm}
  \caption{\textbf{\methodname enables computer use agents self-evolving 
 in novel environments} by autonomously exploring and learning from their own experiences without human intervention. The specialist-to-generalist training strategy further enhances the development of a strong generalist agent.}
  \vspace{-6mm}
  \label{fig:teaser}
\end{figure*}

% To address this challenge, we propose \methodname, an agentic self-evolving framework in which Computer Use Agents (CUAs) are exposed to previously unfamiliar software environments and engage in autonomous exploration and experiential learning, as illustrated in Fig.~\ref{fig:teaser}. Enabling such self-evolution requires overcoming two key challenges: (1) generating executable tasks within unfamiliar software environments, and (2) accurately judging task success and precisely identifying the failure steps. To this end, we introduce a World State Model for environmental state captioning and step-wise trajectory assessment, along with a Curriculum Generator supported by a continuously updated software guidebook memory to produce increasingly diverse and challenging tasks. The agent's policy is trained via autonomous reinforcement learning, combining adversarial imitation of failure actions with Group Relative Policy Optimization (GRPO) on successful ones.

To address this challenge, we propose \methodname, an agentic self-evolving framework in which Computer Use Agents (CUAs) are exposed to previously unfamiliar software environments and engage in autonomous exploration and experiential learning, as illustrated in Fig.~\ref{fig:teaser}. Enabling such self-evolution requires addressing two key challenges: (1) generating executable tasks within unfamiliar software environments, and (2) accurately assessing task success and precisely identifying the step at which failure occurs.
To this end, we introduce a \textbf{World State Model} for environmental state captioning and step-wise trajectory assessment, together with a \textbf{Curriculum Generator} powered by a continuously updated software guidebook memory to generate increasingly diverse and challenging tasks, thereby establishing a curriculum learning paradigm. The agent's policy is optimized through experiential learning from both failures and successes, combining adversarial imitation of failure actions and Group Relative Policy Optimization (GRPO) on successful ones.

Given the critical role of reward accuracy, we conduct extensive evaluations and observe that existing reward models of computer use tasks fall short in terms of judgment precision and reward density. Leveraging the enhanced long-context processing capabilities of advanced LVLMs, we input the agent’s full trajectory of states into the reward model and fine-tune a reward model, \judge, using Qwen2.5-VL~\cite{bai2025qwen2}, substantially narrowing the gap with commercial models such as GPT-4o~\cite{gpt4} with +7.5\% improvement in precision compared to baseline model in evaluating CUAs' trajectories on AgentRewardBench~\cite{lu2025agentrewardbench}, enable \judge to provide high quality step level reward signals in self-evolving agentic system.

% Besides, \methodname supports the self-evolution of computer-use agents, whether as specialists tailored to a single software or as generalists capable of operating across multiple software platforms. However, we find that training generalist agents directly yields suboptimal results. To tackle the problem, we adopt a specialist-to-generalist training strategy, which leads to superior overall performance compared to training generalist agents alone.

% Besides, \methodname enables agents to evolve as either single-software specialists or a multi-software generalist. To tackle the observation that direct training of a generalist underperforms than specialists, we further introduce a novel specialist-to-generalist training strategy, even outperforming the ability of each specialist on its specialized software application.

Moreover, \methodname enables agents to evolve into either single-software specialists or multi-software generalists. To overcome the limitation that directly training a generalist underperforms compared to specialists, inspired by \cite{zhang2024buildingspecializedgeneralistai}, we introduce a novel specialist-to-generalist training strategy, which even surpasses the performance of individual specialists on their respective software applications.

We perform extensive experiments of \methodname built on UI-TARS~\cite{qin2025uitars} and evaluated on five professional software applications from OSWorld~\cite{xie2024osworld}. \methodname with the specialist-to-generalist strategy significantly improves the UI-TARS~\cite{qin2025uitars} from 11.3\% to 34.5\%. Furthermore, \methodname with the specialist-to-generalist strategy (34.5\%) outperforms both specialist RL (32.2\%) and generalist RL (30.6\%) by a substantial margin, demonstrating the effectiveness of the specialist-to-generalist paradigm. In general, \methodname offers a promising approach for developing more powerful and versatile computer-use agents without human involvement.
% Moreover, our method outperforms previous RL-based  approaches~\cite{bai2024digirl,qi2024webrl} using Generalized Advantage Estimation~\cite{schulman2015high} (21.8\%) 

% Our core contributions are as follows:

% \textbf{(1)} We present a novel framework that enables computer use agents (CUAs) to learn from their own experiences, facilitating specialization in previously unseen software environments.

% \textbf{(2)} We propose \judge, a reward model that processes the full sequence of screenshots from agent interactions and delivers fine-grained, step-level reward signals to guide reinforcement learning. \judge demonstrates high precision on AgentRewardBench~\cite{lu2025agentrewardbench}.

% \textbf{(3)} We conduct experiments across five professional software environments in OSWorld~\cite{xie2024osworld}, where \methodname achieves new state-of-the-art performance, surpassing prior RL-based approaches~\cite{bai2024digirl,qi2024webrl}.

% \textbf{(4)} We introduce a specialization-to-generalization strategy, where specialists trained on individual software are distilled into a generalist model that surpass the ensemble of specialists.

\section{Related Work}
% \paragraph{Agent for Computer Use.} With recent revolution in LLM and LVLMs~\cite{touvron2023llama,grattafiori2024llama,liu2023visual,bai2025qwen2,wang2024qwen2}, processing human level perception and reasoning ability, building computer use agent is also intensively studied~\cite{hu2024agents,hong2024cogagent,cheng2024seeclick,nguyen2024gui,lin2024showui}. These agents either takes only text input from structured text~\cite{qi2024webrl} or more like human, take screenshot and text condition as multi-modal input. Although intensively studied and perform well on in-domain benchmark~\cite{lu2024gui,zheng2024gpt,liu2024visualwebbench,li2025screenspot,cheng2024seeclick}, The computer use agent still fall largely behind human level performance in simulation environment~\cite{xie2024osworld,rawles2024androidworld,koh2024visualwebarena,zhou2023webarena}, as its challenge the multi-dimension ability of LVLMs in grounding, decision making and reasoning with works done breaking this process into different expert models~\cite{gou2024navigating,wan2024omniparser} with agent calloberation~\cite{agashe2024agent,agashe2025agent,liu2023bolaa,zhang2025appagent} through prompt engineering~\cite{yan2023gpt,he2024webvoyager,zhang2024android,wang2023voyager,wu2024copilot}, However, these training free methods improvements is restricted without fine-tuning. In this work, we dive into the next step of CUA where the pretrained agent is fine-tuned to learn from its own experience and achieves self-evolution on specialized novel software without human annotations.

\paragraph{Agent for Computer Use.} With the recent advances in LLMs and LVLMs~\cite{touvron2023llama,grattafiori2024llama,liu2023visual,bai2025qwen2,wang2024qwen2}, which enable human-level perception and reasoning capabilities, the development of agents for computer use has garnered significant attention~\cite{hu2024agents,hong2024cogagent,cheng2024seeclick,nguyen2024gui,lin2024showui}. These agents either rely solely on structured text inputs~\cite{qi2024webrl,nakano2021webgpt,putta2024agent,lai2024autowebglm,ma2023laser} or, in a more human-like manner, use multi-modal inputs such as screenshots combined with textual conditions~\cite{hong2023cogagent,lin2024showui,wu2024atlas,operator}. Although they have been extensively studied and show strong performance on in-domain benchmarks~\cite{lu2024gui,zheng2024gpt,liu2024visualwebbench,li2025screenspot,cheng2024seeclick}, computer use agents still lag significantly behind human-level performance in simulated environments~\cite{xie2024osworld,rawles2024androidworld,koh2024visualwebarena,zhou2023webarena}. This gap highlights the challenges posed by the multi-dimensional demands on LVLMs, including grounding, decision-making, and reasoning. Some approaches address this by decomposing tasks into specialized expert models~\cite{gou2024navigating,wan2024omniparser} and employing agent collaboration~\cite{agashe2024agent,agashe2025agent,liu2023bolaa,zhang2025appagent} through prompt engineering~\cite{yan2023gpt,he2024webvoyager,zhang2024android,wang2023voyager,wu2024copilot}. However, improvements from these training-free methods remain limited without fine-tuning. In this work, we explore the next phase of computer use agents, where a pretrained agent is fine-tuned to learn from its own experience, enabling self-evolution on novel, specialized software without human annotations.

% \paragraph{Reinforcement Learning for LLM/ LVLMs.} Previous scalable training for LLM/ LVLMs~\cite{touvron2023llama,grattafiori2024llama,liu2023visual,bai2025qwen2,wang2024qwen2} mainly from supervised fine-tuning (SFT)~\cite{liu2023visual,wei2022chain}. Similar to imitation learning in RL, SFT teach model to output labeled desire output. This makes SFT highly dependent on high quality human procedure data. Recently, DeepSeek-R1~\cite{guo2025deepseek} achieve strong reasoning ability through Group Relative Policy Optimization (GRPO)~\cite{shao2024deepseekmath} with verifiable rewards. Previous works~\cite{ouyang2022training,ziegler2019fine,rafailov2023direct} also apply RL to single turn optimization from human feedback. However, in agentic applications like computer use where environment feedback is sparse, where success is achieved with multi-step interactions. it is important to introduce stable step level reward signals. Previous works on RL for agent~\cite{bai2024digirl,qi2024webrl,zhou2024archer,zhai2024fine,carta2023grounding} fine-tune their own critic model for advantage estimation based on output reward model (ORM) finetuned on labeled data or use DPO~\cite{rafailov2023direct} policy updates based on interaction data~\cite{putta2024agent,qin2025uitars}. In this work, we dive into evaluation of different strategies for building the best performing reward model for CUAs and find that full process based analysis provide the most accurate results compared to training specific critic model to perform advantage estimation in~\cite{bai2024digirl,qi2024webrl}.

\paragraph{Reinforcement Learning for LLMs/LVLMs.} Previous scalable training efforts for LLMs and LVLMs~\cite{touvron2023llama,grattafiori2024llama,liu2023visual,bai2025qwen2,wang2024qwen2,xing2025scalecap,sun2024bootstrap3d,sun2024xpromptuniversalincontextimage,ding2025mm} have primarily relied on supervised fine-tuning (SFT)~\cite{liu2023visual,wei2022chain}. Analogous to imitation learning or behavior cloning in reinforcement learning (RL), SFT trains models to produce desired outputs based on labeled data, making it heavily dependent on high-quality human-curated procedures. Recently, DeepSeek-R1~\cite{guo2025deepseek} demonstrated strong reasoning capabilities via Group Relative Policy Optimization (GRPO)~\cite{shao2024deepseekmath} using verifiable rewards. Earlier works~\cite{ouyang2022training,ziegler2019fine,rafailov2023direct} have also employed RL for single-turn optimization from human feedback. However, in agentic scenarios such as computer usage—where feedback is sparse with reward signals often results from multi-step interactions—it becomes crucial to introduce stable, step-level reward signals. Prior RL approaches for agents~\cite{bai2024digirl,qi2024webrl,zhou2024archer,zhai2024fine,carta2023grounding} have fine-tuned their own critic models for advantage estimation~\cite{schulman2015high}, either using output reward models (ORMs) trained on labeled data or adopting Direct Preference Optimization (DPO)~\cite{rafailov2023direct} based on interaction data~\cite{putta2024agent,qin2025uitars}. In this work, we investigate various strategies for constructing high-performing reward models for CUAs and find that full-process-based analysis yields more accurate evaluations with fine-grained reward signals compared to training dedicated critic models for advantage estimation as done in~\cite{bai2024digirl,qi2024webrl} or with filtered behavior cloning~\cite{pan2024autonomous,chen2020bail}.

\section{Methods}
\noindent \textbf{Problem Formulation.} The objective of \methodname is to establish a training pipeline enabling the Computer Use Agent (CUA) to autonomously explore its environment (Sec.~\ref{sec:autonomous_exploration_pipeline}) and progressively self-evolve on novel software applications via reinforcement learning from experience (Sec.~\ref{sec:rl}). Specifically, the \methodname pipeline comprises three primary components: an Actor Model $\pi$ performing exploratory actions to accomplish these tasks, and a World State Model $\mathcal{M}_{state}$ describing the current environment state and evaluating the success or failure of executed actions, and a Curriculum Generator $\mathcal{M}_{task}$ that continuously proposes more diverse and challenging exploration tasks:

\textbf{(1) Actor Model $\pi$:} The policy $\pi(a|s_t, I)$ defines the probability of taking action $a$ at time step $t$, conditioned on the current state $s_t$ and the overall task instruction $I$.

\textbf{(2) World State Model $\mathcal{M}_{state}$:} This component is a fine-tuned Large Vision-Language Model (LVLM) responsible for providing detailed descriptions of environment states. It also evaluates each step of the trajectory executed by the Actor Model $\pi$, producing trajectory judgement \( \mathcal{J} \) which indicates whether the task has been successfully completed. Joint training with state change captioning \( \mathcal{C} \) of the software GUI has been shown to enhance judgment accuracy, as shown in Table~\ref{tab:judge_eval}.

\textbf{(3) Curriculum Generator $\mathcal{M}_{task}$:} This component utilizes a powerful Large Language Model (LLM) to automatically generate novel exploration tasks. It also maintains and updates a software guidebook $U$ based on the state change captioning \( \mathcal{C} \) and the trajectory judgement \( \mathcal{J} \) provided by $\mathcal{M}_{state}$ during interactions. The gradually enriched guidebook $U$ enables $\mathcal{M}_{task}$ to progressively generate increasingly diverse and challenging tasks in a curriculum learning fashion.

\methodname can be applied to enable the self-evolution of a computer-use agent, either as a specialist for a single software or as a generalist across multiple software. However, we observe that direct training for generalist agents is suboptimal. We introduce a specialist-to-generalist training strategy, which achieves even better overall performance than training multiple generalist agents, as discussed in Sec.~\ref{sec:sg}.

% {\color{red}{why use llm rather than lvlm as Curriculum Generator should be discussed.}}

\subsection{Autonomous Exploration with Self-evolving Curriculum}
\label{sec:autonomous_exploration_pipeline}

\begin{figure*}[t]
  \centering
  \includegraphics[width=1.0\linewidth]{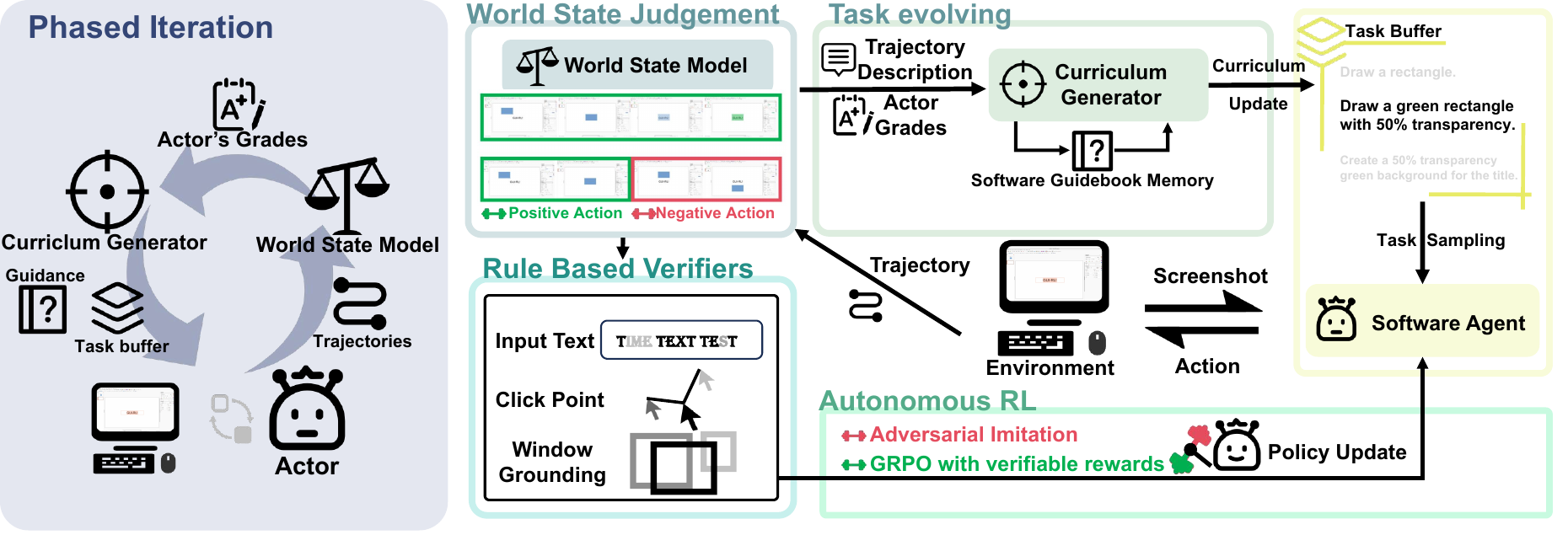}
  \vspace{-6mm}
  \caption{\textbf{\methodname autonomous exploration and experiential learning pipeline.} Guided by tasks generated by the Curriculum Generator, the Actor Model is updated according to step-level rewards from the World State Model through verifiable reward functions tailored for different action types.}
  \vspace{-6mm}
  \label{fig:data_pipeline}
\end{figure*}

% Autonomous exploration is essential for enabling the Computer Use Agent (CUA) to become proficient with a novel software application. Two critical challenges need to be tackled to enable autonomous exploration, which are specifically to generate tasks to be executed and judge whether the task is completed successfully, and if the task fails, in which step is wrong. Leveraging the introduces novel components World State Model $\mathcal{M}_{state}$ and Curriculum Generator $\mathcal{M}_{task}$, we formulate a task self-evolving curriculum paradimg to generating increasingly more diverse and challenging tasks to be explored. 
% Autonomous exploration is crucial for enabling the Computer Use Agent (CUA) to gain proficiency in novel software applications, which are previously unseen or poorly understood. Two core challenges must be addressed: (1) generating executable tasks within unfamiliar software environments, and (2) evaluating whether a task has been completed successfully, and if not, identifying the specific step at which failure occurred. To address these challenges, we introduce two novel components: the World State Model $\mathcal{M}{\text{state}}$ and the Curriculum Generator $\mathcal{M}{\text{task}}$. Together, these two components support a \textbf{self-evolving curriculum paradigm} that generates increasingly diverse and challenging tasks for autonomous exploration. 
Autonomous exploration is essential for enabling the Computer Use Agent (CUA) to develop proficiency in novel software applications that are previously unseen or poorly understood. This process involves addressing two key challenges: (1) generating executable tasks within unfamiliar software environments, and (2) evaluating task completion success and pinpointing the specific step at which failure occurs. To tackle these challenges, we introduce two novel components: the World State Model $\mathcal{M}_{\text{state}}$ and the Curriculum Generator $\mathcal{M}_{\text{task}}$. These components jointly support a \textbf{self-evolving curriculum paradigm}, which facilitates the autonomous generation of increasingly diverse and challenging tasks.

The \textbf{self-evolving curriculum paradigm} pipeline is structured into $P$ sequential phases. Before the first phase, a set of initial tasks targeting basic GUI operations is generated (details provided in Sup.~\ref{sup:data_pipe}). In each phase, these tasks are executed and step-wise evaluated. The resulting judgments and descriptions of the exploration trajectories are fed into the Curriculum Generator $\mathcal{M}_{\text{task}}$, which updates a self-maintained software guidebook $U$. Leveraging this updated guidebook and the current capabilities of the CUA, the generator then produces more diverse and challenging tasks for subsequent phases. The following outlines each step of the process in detail:

\textbf{(1) Task initiation:} The initial state of the unfamiliar software is provided, typically in the form of screenshots of its basic GUI interface. The World State Model $\mathcal{M}_{\text{state}}$ performs dense captioning of the GUI elements, including button detection and OCR-based recognition. These detailed captions are passed to the Curriculum Generator $\mathcal{M}_{\text{task}}$, which generates an initial set of task instructions \( \mathcal{I}_0 = \{I_0^{(1)}, I_0^{(2)}, \cdots \} \) along with an initial software guidebook \( U_0 \) for the software.

\textbf{(2) World state judgment:} In the $p$-th phase of \textit{Auto Exploration}, the Actor Model $\pi_p$ executes tasks based on the instructions in \( \mathcal{I}_{p} \). Each execution is evaluated by the World State Model $\mathcal{M}_{\text{state}}$, which provides feedback \( \mathcal{J}_{p} = \{J_{p}^{(1)}, J_{p}^{(2)}, \cdots \} \) for each step within the operation trajectory. In addition, it generates a detailed description of GUI state changes based on captured screenshots, denoted as \( \mathcal{C}_{p} \).

\textbf{(3) Task self-evolving:} Based on the outcomes $\mathcal{J}_{p}$ and $\mathcal{C}_{p}$, the Curriculum Generator $\mathcal{M}_{\text{task}}$ produces a more challenging task set \( \mathcal{I}_{p+1} \) and expands the agent's knowledge boundary by updating the software guidebook to \( U_{p+1} \). The detailed prompting process is illustrated in Fig.~\ref{fig:task_buffer_prompt}. This iterative update can be formalized as:
\begin{equation}
U_{p+1}, \mathcal{I}_{p+1} = \mathcal{M}_{\text{task}}(U_p, \mathcal{I}_p, \mathcal{J}_p, \mathcal{C}_p)
\end{equation}
Here, $U_{p+1}$ serves as a more comprehensive software guidebook memory, while $\mathcal{I}_{p+1}$ represents a more challenging task set tailored to the current capabilities of the Actor Model $\pi_p$. Examples of $\mathcal{I}_p$ are provided in Fig.~\ref{fig:results}, where the Actor Model $\pi$ demonstrates curriculum learning by handling increasingly complex tasks across different phases $p$. Illustrations of $U_p$ across various software applications are provided in Sup.~\ref{sup:usage_manual}. Comparison with previous methods~\cite{murty2025nnetnavunsupervisedlearningbrowser,murty2024bagelbootstrappingagentsguiding,sun2024genesis} on task generation are detailed in Sup.\ref{sup:comparision_task_gen}

\textbf{(4) Autonomous RL Training:} Through iterative reinforcement learning, the Actor Model $\pi_p$ is updated based on its execution of the instruction set~\( \mathcal{I}_p \), guided by evaluation feedback \( \mathcal{J}_p \) and a set of action-specific verifiable functions \( \mathcal{R}_{\text{verifer}} \). The resulting policy $\pi_{p+1}$ is then used as the actor in the subsequent phase. Further details are provided in Sec.~\ref{sec:rl}.

\subsection{Reinforcement Learning from Experience}
\label{sec:rl}

The World State Model $\mathcal{M}_{state}$ provides step-level reward signals for reinforcement learning. Unlike previous reward models for CUA~\cite{qi2024webrl,bai2024digirl,putta2024agent,pan2024autonomous,lu2025agentrewardbench}, 
%which were developed under the constraints of less capable LVLMs, 
our $\mathcal{M}_{state}$ model takes the entire trajectory of states and actions, $\mathcal{H} = \{(s_0, a_0), (s_1, a_1), \ldots\}$, as input. It classifies each action $a$ as either $a_F$ or $a_T$, where $a_F$ indicates an incorrect action leading to failure or redundant loops, and $a_T$ represents a correct action that contributes to successful progression without redundancy. The curated prompt used for judgment is depicted in Fig.~\ref{fig:os_prompt}. For historical states that result in $a_T$, we encourage CUA to reinforce these actions through verifiable rewards defined by a set of functions \( \mathcal{R}_{\text{verifer}} = \{ r_{dist}\}\). Conversely, for states leading to $a_F$, we penalize them using negative KL divergence with adversarial imitation.

\noindent \textbf{Adversarial Imitation for Failure Action Punishment.}  
To explicitly encourage the policy to diverge from failure-inducing behaviors, we employ a contrastive log-ratio loss based on a reference failure action \( a_F \). This objective encourages the policy to sample actions \( a \) that minimize alignment with the failure action \( a_F \):
\begin{equation}
\mathcal{L_{\text{AI}}}(\pi_\theta) = \mathbb{E}_\nu \left[ - \log \frac{\pi_\theta(a \mid s, I)}{\pi_{\text{ref}}(a_F \mid s, I)} \right]
\end{equation}

This formulation serves as an adversarial imitation signal. By maximizing divergence from this distribution, the agent is trained to explore alternative action distributions that deviate from those leading to failure, particularly in complex GUI interaction scenarios. Notably, this loss shares a similar form with DPO~\cite{rafailov2023direct} but only the negative part.

\noindent \textbf{Verifiable Rewards for Correct Action Encouragement.}  
To more effectively guide the policy toward correct actions $a_T$, we adopt Reinforcement Learning with Verifiable Rewards (RLVR)~\cite{guo2025deepseek,shao2024deepseekmath}, which has recently shown success in enhancing language models on tasks with objectively verifiable answers, such as mathematics~\cite{shao2024deepseekmath}, and more recently, counting and grounding in the vision-language domain~\cite{liu2025visual,shen2025vlm,meng2025mm}. After labeling the correct step $(s, a_T)$ using the World State Model, we apply Group Relative Policy Optimization (GRPO), computing the relative advantage of each response based on its reward:

\begin{equation}
A^{(i)} = 
 \frac{r^{(i)} - \text{mean}(\{r^{(j)}\}_{j=1}^{G})}{\text{std}(\{r^{(j)}\}_{j=1}^{G})}, \quad i=1,\cdots,G.
\end{equation}
As we design distinct reward signals for different action types, we define the reward function between a predicted action $a$ and the ground-truth action $a_T$ as:
\begin{equation}
r^{(i)} = r(a^{(i)}, a_T) = \mathbb{I}\left(\text{type}(a^{(i)}) = \text{type}(a_T)\right) + r_{\text{dist}}(a^{(i)}, a_T),
\end{equation}
where $\mathbb{I}(\cdot)$ is the indicator function that returns 1 if the predicted action and ground-truth action are of the same type, and 0 otherwise. The distance-based reward term $r_{\text{dist}}(a^{(i)}, a_T)$ is defined according to the specific action type: for \texttt{click} actions, it is computed based on the normalized L1 distance between the clicked coordinates; for \texttt{drag} and \texttt{select} actions, it is computed using the Intersection over Union (IoU) between the predicted and ground-truth bounding boxes; and for \texttt{type} actions, it is determined by the character-level BLEU score between the predicted and ground-truth text. All $r_{\text{dist}}$ rewards are normalized to the range $[0, 1]$ to ensure consistency across different action types. A comprehensive list of $r_{\text{dist}}(a^{(i)}, a_T)$ definitions for various action types is provided in Tab.~\ref{tab:reward_list}.  The final loss of GRPO is directly adopted from~\cite{shao2024deepseekmath}:

\begin{align}
\mathcal{L}_{\text{GRPO}}(\pi_\theta) &= 
- \mathbb{E}_{(s, I) \sim \mathcal{D}, \{a^{(i)}\}_{i=1}^G \sim \pi_{\text{ref}}(\cdot \mid s, I)}
\\\Bigg[ \frac{1}{G} \sum_{i=1}^G \frac{1}{|a^{(i)}|}& \sum_{t=1}^{|a^{(i)}|}
\Big\{
\min\Big(
r_t^{(i)}(\theta) A^{(i)}, 
\text{clip}(r_t^{(i)}(\theta), 1 - \epsilon, 1 + \epsilon) A^{(i)}
\Big)
- \beta \, D_{\text{KL}}^{(i,t)}(\pi_\theta \| \pi_{\text{ref}})
\Big\}\Bigg],\nonumber
\end{align}

% where

\begin{align*}
    \text{where} \quad
    r^{i,t}(\theta) = \frac{\pi_{\theta}(a^{(i)}|s, I)}{\pi_{\theta_\text{ref}}(a^{(i)}|s, I)} 
    \text{ and }
    D_\text{KL}^{i,t}(\pi_\theta, \pi_\text{ref})=
    \frac{\pi_\text{ref}(a^{(i)}|s, I)}{\pi_{\theta}(a^{(i)}|s, I)} - 1 - \log \frac{\pi_\text{ref}(a^{(i)}|s, I)}{\pi_{\theta}(a^{(i)}|s, I)}.
\end{align*}

Similar to~\cite{shao2024deepseekmath,guo2025deepseek}, advantage $A$ is weighted on the whole reasoning token logits to encourage free form thinking for performing action and planning. 

The final training loss is defined as a weighted combination of positive and negative action samples, i.e., correct actions $a_T$ and incorrect actions $a_F$: 
$\mathcal{L}(\pi(\theta)) = \mathcal{L}_{\text{GRPO}} + \gamma \mathcal{L}_{\text{AI}}$. 
We set $\gamma = 0.2$ during training, and the rationale for this choice is discussed in the ablation study presented in Sup.~\ref{sup:loss_factor}.

This strategy is shown to be more effective in Sec.~\ref{sec:eval_cua} compared to Generalized Advantage Estimation (GAE)~\cite{schulman2015high}-based RL methods~\cite{qi2024webrl,bai2024digirl}, as the more powerful reward model $\mathcal{M}_{state}$ provides accurate step-level reward signals by leveraging the entire episode trajectory $\mathcal{H}$ from a global perspective.

\subsection{From Specialist to Generalist.}
\label{sec:sg}
Achieving a generalist agent capable of operating across multiple software platforms is an ambitious and valuable goal. We first attempted to train such a generalist directly using the proposed \methodname framework across all software environments. However, this approach led to suboptimal performance compared to specialized agents, as the actor struggled to learn effectively in the multi-software environment.

We thus introduce a specialist to generalist strategy, as illustrated in Fig.~\ref{fig:teaser}. Specifically, we first train software-specialized agents via \methodname on individual environments, allowing each to master a specific application. These specialists are then distilled into a single generalist model through supervised fine-tuning (SFT) on synthesized successful trajectories. Finally, the generalist is refined via \methodname on multiple software. This generalist, now equipped with better reasoning, planning abilities, and software-specific commonsense, achieves significantly improved performance, outperforming both the \methodname via direct general RL and the performance combination of multi-specialists as in Table~\ref{tab:osworld_res}.

\section{Experiments}

\subsection{Benchmark of Reward Model for computer use agent.}
\label{sec:gui_judge}
% Provide CUAs with reliable reward signal is crucial for the agentic system (actor CUA and judge model) to achieve self-evolution on unfamiliar software. Recent work AgentRewardBench~\cite{lu2025agentrewardbench} proposes to evaluate the precision of judges from reward model on web tasks with trajectories from diversified agents. Beyond AgentRewardBench~\cite{lu2025agentrewardbench} in web tasks, we also extend the evaluation to diversified PC softwares. We use all 339 feasible tasks from OSWorld~\cite{xie2024osworld} with rule-based evaluation. We sample trajectories from UI-TARS~\cite{qin2025uitars} 7B/ 72B and use rule-based evaluation as ground truth label for success/ failure and compute confusion matrix with different reward models' predictions.

Providing CUA agents with reliable reward signals is crucial for enabling self-evolution in agentic systems, consisting of an actor (CUA) and a judge model, especially when interacting with unfamiliar software environments. Recent work, AgentRewardBench~\cite{lu2025agentrewardbench}, proposes to evaluate the precision of reward models by assessing the accuracy of judge predictions on web-based tasks using trajectories from diverse agents. 
Building upon AgentRewardBench~\cite{lu2025agentrewardbench}, we further extend the evaluation beyond web tasks to a broader set of PC software environments. Specifically, we evaluate on all 339 feasible tasks from OSWorld~\cite{xie2024osworld}, using rule-based criteria for determining success or failure. Trajectories are sampled from UI-TARS~\cite{qin2025uitars} and Gemini-2.5-Pro~\cite{google2025gemini25preview}, and rule-based evaluation is used as ground-truth supervision. We then compute the confusion matrix by comparing the predictions of different reward models against these labels.

% The judge strategy in AgentRewardBench~\cite{lu2025agentrewardbench} are all based on the final state with history actions. However, it is nature for the judge to take the whole trajectory into consideration to judge the successness instead of only the last state. For example, If you want the agent to buy a ticket to Landon, The final state of "Your flight ticket has been successfully booked." may not tell whether the agent pick the correct date and time, leading to compromised judge precision. However, we find it not the case for current open-sourced LVLMs. As depicted in Fig.\ref{fig:process_curve}, inputting more history screenshots lead to huge drop on Average Precision (AP) for Qwen2.5-VL~\cite{bai2025qwen2}, which diveraged with GPT-4o~\cite{hurst2024gpt} on the same curated prompt, we attribute this to the inadequte pretraining of Qwen2.5-VL~\cite{bai2025qwen2} on sequencial high resolution screenshot images approaching context length limits. We find this can be easily fixed by distill model to provide step by step screenshot analysis towards final judgement and name this model \judge. The training process of \judge is detailed in Sup.\ref{sec:gui_judge_train} with training data from 0.86K GPT-4o~\cite{hurst2024gpt} on Chrome software in OSWorld~\cite{xie2024osworld} to provide different-sourced data from AgentRewardBench~\cite{lu2025agentrewardbench} and other professional software in OSWorld.

The judge strategy in AgentRewardBench~\cite{lu2025agentrewardbench} relies solely on the final state and the associated action history. However, it is more natural and reliable for a judge model to consider the entire trajectory when assessing task success, rather than focusing only on the final state. For example, consider the task of booking a flight to London. A final state message such as "Your flight ticket has been successfully booked." does not confirm whether the correct date and time were selected, which can lead to compromised judgment accuracy.

\begin{wrapfigure}{r}{0.5\linewidth} % r=右侧，l=左侧
  \vspace{-10pt} % 根据需要微调
  \includegraphics[width=\linewidth]{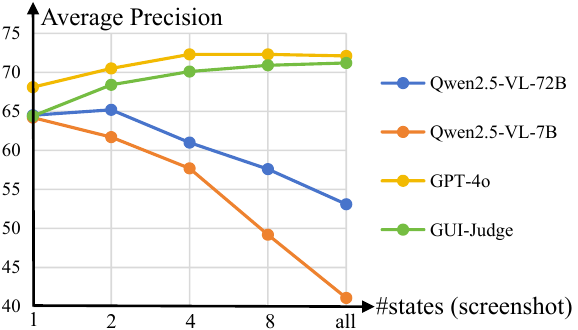}
  \vspace{-6mm}
  \caption{\textbf{The Average Precision on AgentRewardBench~\cite{lu2025agentrewardbench}}, where GUI-Judge exhibits an improvement in AP as the number of input middle states increases, showing a similar trend to that of the closed sourced GPT-4o~\cite{hurst2024gpt} when compared with its base model.}
  \vspace{-3mm}
  \label{fig:process_curve}
  \vspace{-10pt} % 根据需要微调
\end{wrapfigure}

However, we observe that current open-sourced LVLMs do not perform well under this more holistic evaluation strategy. As shown in Fig.~\ref{fig:process_curve}, feeding additional historical screenshots into Qwen2.5-VL~\cite{bai2025qwen2} significantly degrades its Average Precision (AP), diverging notably from GPT-4o~\cite{hurst2024gpt} on the same curated prompt detailed in Fig.\ref{fig:web_prompt}. We attribute this performance drop to the insufficient pretraining of Qwen2.5-VL on long sequences of high-resolution screenshots, which likely pushes it toward the limits of its 32K context length.

\begin{table}[t]
  \centering
    \caption{\textbf{Precision and Negative Predictive Value (NPV)} on AgentReardBench~\cite{lu2025agentrewardbench} and OSWorld~\cite{xie2024osworld} with last screenshot only (LS) or entire process screenshots (ES) as input. \judge closes the gap with commercial model supporting full process high resolution screenshots as input. The co-training with screenshot change description (CD) improves judgment precision.}
    \setlength\tabcolsep{5.4pt}
    \scalebox{0.94}{
    \begin{tabular}{lccccccc}
    \toprule
    \multicolumn{1}{l}{\multirow{2}[3]{*}{\textbf{Model}}} & \multicolumn{1}{c}{\multirow{2}[3]{*}{\textbf{Input}}} & \multicolumn{2}{c}{\textbf{AgentRewardBench}} & \multicolumn{2}{c}{\textbf{OS-World-Full}} & \multicolumn{2}{c}{\textbf{Prof/Office}} \\
    \multicolumn{1}{r}{} & \multicolumn{1}{c}{} & \multicolumn{1}{c}{Precision} & \multicolumn{1}{c}{NPV} & \multicolumn{1}{c}{Precision} & \multicolumn{1}{c}{NPV} & \multicolumn{1}{c}{Precision} & \multicolumn{1}{c}{NPV} \\
    \midrule
    \multirow{2}[1]{*}{GPT-4o~\cite{hurst2024gpt}} & LS    & 68.1  & 92.3  & 46.3  & 88.2  & 40.5  & 81.0  \\
          & ES    & 72.1  & 92.2  & 74.6  & 95.2  & 70.4  & 85.3  \\
    \midrule
    \multirow{2}[1]{*}{Qwen2.5-VL-72B~\cite{bai2025qwen2}} & LS    & 64.5  & 94.2  & 41.5  & 86.9  & 31.7  & 78.7  \\
          & ES    & 26.2  & 83.0  & 26.8  & 83.0  & 25.6  & 76.6  \\
    \midrule
    \multirow{2}[1]{*}{Qwen2.5-VL-7B~\cite{bai2025qwen2}} & LS    & 64.1  & 90.3  & 37.3  & 85.2  & 31.8  & 79.0  \\
          & ES   & 25.4  & 83.8  & 20.0  & 81.7  & 23.5  & 76.0  \\
    \midrule
    \judge (w/o CD) & ES    & 69.1  & 88.5  & 71.1  & 88.4  & 65.0  & 81.1  \\
    \judge (w/ CD) & ES    & 71.6  & 91.2  & 73.9  & 90.5  & 69.3  & 82.0  \\
    % \judge & ES    & 71.6  & 91.2  & 73.9  & 90.5  & 69.3  & 82.0  \\
    \bottomrule
    \end{tabular}%
    }
    \vspace{-4mm}
  \label{tab:judge_eval}%
\end{table}%

% To address this issue, we propose a distilled model, referred to as \judge, which performs step-by-step screenshot analysis to arrive at the final judgment. The training process for \judge is described in Sup.~\ref{sec:gui_judge_train}, using a dataset of 0.86K GPT-4o~\cite{hurst2024gpt}-generated evaluations on trajectories with GUI change dense-descriptions on Chrome within OSWorld~\cite{xie2024osworld}. This dataset provides a distinct source from AgentRewardBench~\cite{lu2025agentrewardbench}, and also includes examples from various professional software within OSWorld.

To address this issue, we propose a distilled model based on Qwen2.5-VL-7B, referred to as \judge, which conducts step-by-step screenshot analysis to produce final judgments. The training process for \judge is detailed in Sup.~\ref{sup:gui_judge_train}, using a dataset of 0.86K GPT-4o~\cite{hurst2024gpt} generated evaluations on trajectories with dense GUI change descriptions, exclusively from Chrome within the OSWorld~\cite{xie2024osworld} environment. Despite being trained solely on Chrome data, \judge exhibits strong generalization to both other professional software in OSWorld and the external AgentRewardBench~\cite{lu2025agentrewardbench} benchmark. This demonstrates that the model learns transferable judgment patterns rather than overfitting to the specifics of a single application, thanks to the diversity and quality of step-level annotations in the training data.

% We test \judge and our entire process screenshots conditioned strategy on full AgentRewardBench~\cite{lu2025agentrewardbench} and agents' trajectories from OSWorld~\cite{xie2024osworld}. As reported in Tab.\ref{tab:judge_eval} and ablated in Fig.\ref{fig:process_curve}. \judge achieve SOTA open-souced performance, closing the gap with GPT-4o~\cite{hurst2024gpt} and more importantly, demonstrate similar trend of GPT-4o~\cite{hurst2024gpt} when provided with history screenshots. Dispide trained on small amount of data, we fully incentivize the model to understand the sequencial relationship between history images and provide step by step check to conclude the final judgement. \judge serves as the foundation reward model and provide reliable step level reward signal. Based on the design strategy of evolution of an agentic system including both the actor agent and reward judge. we choose not to use API call from GPT-4o~\cite{hurst2024gpt} to provide judgments.

We evaluate \judge and our full-process screenshot-conditioned strategy on AgentRewardBench~\cite{lu2025agentrewardbench}, as well as on agent trajectories from OSWorld~\cite{xie2024osworld}. As shown in Tab.~\ref{tab:judge_eval} and further analyzed in Fig.~\ref{fig:process_curve}, \judge achieves state-of-the-art performance among open-sourced models, significantly narrowing the gap with GPT-4o~\cite{hurst2024gpt}. More importantly, it exhibits a similar performance trend to GPT-4o when conditioned on historical screenshots. Despite being trained on a relatively small dataset, \judge is explicitly encouraged to capture the sequential dependencies among historical screenshots and to perform step-by-step reasoning for final judgment. Serving as our foundation reward model, \judge provides reliable, step-level reward signals that support downstream policy learning. In line with our agentic system design—which emphasizes the evolution of the actor agent with full open-sourced models—we intentionally avoid relying on GPT-4o~\cite{hurst2024gpt} API calls for judgment during training and inference (also due to inefficiency). More details of \judge is supplied in Sup.\ref{sup:gui_judge}.

% \begin{wrapfigure}{r}{0.35\textwidth}
% \centering
% \vspace{-2.2em}
% \includegraphics[width=1.0\linewidth,trim={0 5 0 20},clip]{figures/process_curve.pdf}\vspace{-1.6em}
% \caption{User study.}\vspace{-1.5em}
% \label{fig:user_study}
% \end{wrapfigure}

% \begin{figure}[h]
%   \begin{minipage}{0.50\linewidth}
%     \hfill % push to right side
%     \includegraphics[width=\linewidth]{figures/process_curve.pdf}
%     \caption{\textbf{The Average Precision on AgentRewardBench~\cite{lu2025agentrewardbench}}, where GUI-Judge exhibits an improvement in AP as the number of input middle states increases, showing a similar trend to that of the closed sourced GPT-4o~\cite{hurst2024gpt} when compared with its base model.}
%     \label{fig:process_curve}
%   \end{minipage}
% \end{figure}

\subsection{Self evolution of GUI Agents}
\label{sec:eval_cua}

\begin{table}[t]
\centering
\caption{\textbf{Success Rate (SR) on OSWorld~\cite{xie2024osworld}}. \methodname demonstrates strong performance after reinforcement learning from experience. In addition to evolving on separate software, a new General Model achieves better performance after another iteration of \methodname. *Indicates specialist agents trained separately for each software with ensembled results. All results are averaged over three runs.}
\setlength\tabcolsep{8.1pt}
\scalebox{0.91}{
\begin{tabular}{lcccccc}
\toprule
\textbf{Model} & \textbf{VScode} & \textbf{GIMP} & \textbf{Impress} & \textbf{VLC} & \textbf{Writer} & \textbf{Overall} \\
\midrule
Human Performance & 73.9 & 73.1 & 80.9 & 70.6 & 73.9 & 74.5 \\
\midrule
GPT-4o~\cite{hurst2024gpt} & 4.35 & 3.85 & 6.77 & 16.1 & 4.35 & 7.08 \\
GPT-4V~\cite{gpt4} & 0.00 & 7.69 & 2.52 & 18.3 & 4.35 & 6.59 \\
Gemini-Pro-1.5~\cite{team2023gemini} & 0.00 & 11.5 & 13.2 & 6.53 & 8.71 & 7.99 \\
Claude3.7 Sonnet~\cite{anthropic2025claude37systemcard} & 18.8 & 24.4 & 10.6 & 27.5 & 17.4 & 19.7 \\
Gemini-Pro-2.5~\cite{google2025gemini25preview} & 21.7 & 26.9 & 9.92 & 25.5 & 24.6 & 21.7 \\
\midrule
UI-TARS-7B-DPO~\cite{lu2024gui} & 13.0 & 23.1 & 4.26 & 11.8 & 4.35 & 11.3 \\
UI-TARS-72B-DPO~\cite{lu2024gui} & 18.8 & 25.6 & 6.38 & 15.7 & 8.70 & 15.0 \\
\midrule
DigiRL~\cite{bai2024digirl} (Specialized RL)* & 21.7 & 32.1 & 12.8 & 23.5 & 18.8 & 21.8 \\
WebRL~\cite{qi2024webrl} (Specialized RL)* & 27.5 & 29.5 & 10.6 & 25.5 & 15.9 & 21.8 \\
\methodname (Specialized RL)* & \underline{37.7} & \underline{38.5} & \underline{22.0} & \underline{33.3} & \underline{29.0} & \underline{32.2} \\
\midrule
DigiRL~\cite{bai2024digirl} (General RL) & 21.7 & 35.9 & 12.1	& 19.6 & 15.9 & 21.0 \\
WebRL~\cite{qi2024webrl} (General RL) & 20.3 & 32.5 & 9.93 & 21.6 & 14.5 & 19.6 \\
\methodname (General RL) & 36.2 & 39.7 & 19.9 & 31.4 & 26.1 & 30.6 \\
\midrule
\methodname (General SFT) & 30.4 & 37.2 & 18.4 & 31.9 & 20.3 & 27.9 \\
\methodname (Specialist-to-Generalist) & \textbf{40.5} & \textbf{42.3} & \textbf{22.7} & \textbf{35.3} & \textbf{31.8} & \textbf{34.5} \\
% General Model + \methodname & 46.4 & 42.3 & 24.8 & 39.2 & 34.8 & 37.5 \\
\bottomrule
\end{tabular}
}
\vspace{-2mm}
\label{tab:osworld_res}
\end{table}

\begin{figure*}[h]
  \centering
  \includegraphics[width=1.0\linewidth]{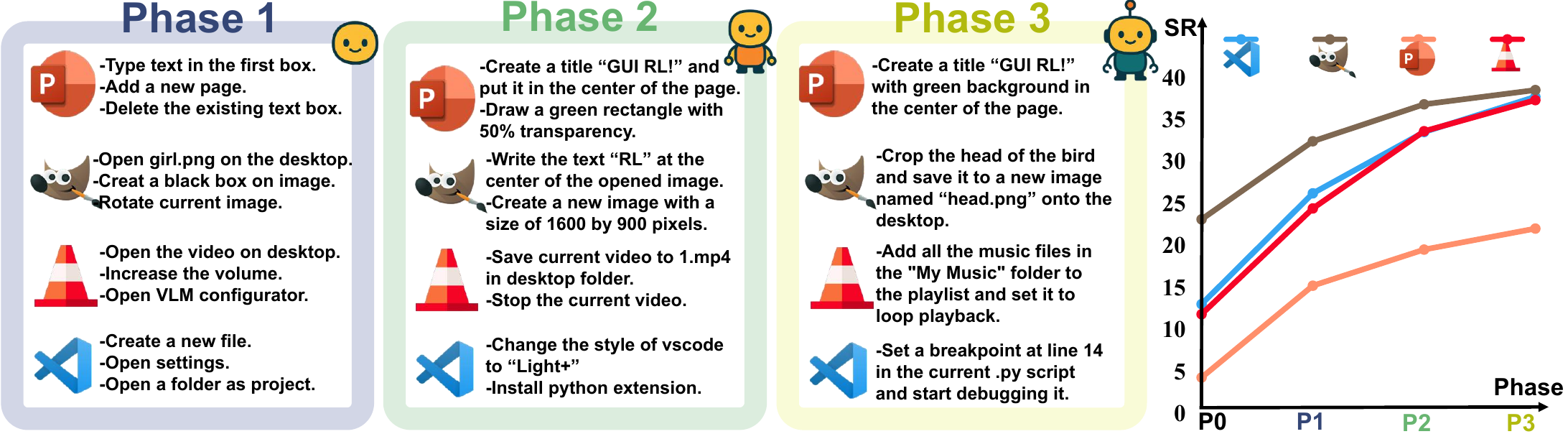}
  \vspace{-6mm}
  \caption{\textbf{Self-evolved task instructions and success rate (SR) curves across different software.} Tasks are progressively upgraded by the Curriculum Generator without human intervention, based on the evolving capabilities of the Actor Model at different training phases.}
  \vspace{-4mm}
  \label{fig:results}
\end{figure*}

\paragraph{Models Before Self-Evolution.}  
Our self-evolving system is initialized with three locally deployed models: UI-TARS-7B-DPO~\cite{qin2025uitars} as the Actor Model, \judge as the step-level reward model, and Qwen2.5-72B~\cite{yang2024qwen2} as the Curriculum Generator for task evolution with software guidebook memory. We conduct experiments on five professional and office-related software applications from OSWorld~\cite{xie2024osworld}. As shown in Tab.~\ref{tab:osworld_res}, the initial actor agent demonstrates limited performance on these software environments, achieving an average success rate of 11.3\% only.

\paragraph{Evolution Process Details.}  
At beginning, we provide \judge with the initial GUI state of the novel software. The Curriculum Generator then generates the first software guidebook and a set of basic tasks (illustrated in Fig.\ref{fig:data_pipeline}). This yields an initial instruction set $\mathcal{I}_0$, averaging 150.2 instructions, which are executed by the Actor Model. The resulting trajectories are evaluated by \judge and parsed into an average of 1361.5 multi-turn conversation pairs (detailed statistics are in Sup.\ref{sup:data_stats}).
We then perform reinforcement fine-tuning (RFT) following the methodology described in Sec.~\ref{sec:rl}. Training is conducted for 1k iterations on 8 NVIDIA A100 80GB GPUs, with $G = 8$, a batch size of 16, and a learning rate of $2 \times 10^{-5}$, scheduled via cosine decay. This evolution process is repeated iteratively for three phases using the same training configuration.

\paragraph{Specialist Evaluation.}  
For a fair comparison with previous reinforcement learning methods~\cite{bai2024digirl,qi2024webrl}, we adapt their training strategies to the UI-TARS~\cite{qin2025uitars} model. Specifically, we initialize the actor agent from UI-TARS-7B-DPO and, instead of providing step-level reward signals, We evaluate its executed trajectories with binary success or failure outcomes using \judge.
A separate critic model is also initialized from UI-TARS-7B-DPO, with additional random initialized MLP layers taking the LLM's hidden states as input to regress value predictions. This critic is trained to perform advantage estimation based on Generalized Advantage Estimation (GAE)~\cite{schulman2015high}. The loss functions follow the same configurations as in~\cite{bai2024digirl,qi2024webrl}. Both the critic and the actor agent are trained iteratively using the same phased reinforcement fine-tuning (RFT) process, where the Curriculum Generator continually generates new curriculum-style tasks.

As shown in Fig.~\ref{fig:results} and Tab.~\ref{tab:osworld_res}, we train separate actor agents for five different software applications. Our approach, denoted as SEAgent (Specialist), achieves strong performance compared to previous reinforcement learning methods such as DigiRL~\cite{bai2024digirl} and WebRL~\cite{qi2024webrl}.  We attribute this improvement to the use of \judge, which provides fine-grained, step-level reward signals derived from a comprehensive understanding of the full history of states and actions. This contrasts with previous approaches that rely on separate critic models—typically initialized from the actor itself—to estimate advantages from sparse, final success/failure signals. Furthermore, the curriculum of task instructions generated by the Curriculum Generator, as illustrated in Fig.~\ref{fig:results}, validates the effectiveness of our autonomous learning framework. These tasks progress from simple to complex based on the actor’s evolving capabilities, enabling it to gradually specialize in each target software environment. Based on the observed evolution curves, we set the number of training phases to three, as performance gains saturate beyond that point.

\paragraph{From Specialist to Generalist.}  
After training five strong software specialists, we pursue generalization using the methodology described in Sec.~\ref{sec:sg}. Specifically, we collect task instructions generated during each specialist's training phase and use them to prompt the respective specialists for execution. A total of 3.5K successful trajectories, along with their corresponding reasoning traces, are distilled into a new base model (UI-TARS-7B~\cite{qin2025uitars}) via supervised fine-tuning (SFT). This distilled model is then further optimized through reinforcement learning (RL) across all five software environments.

As shown in Tab.~\ref{tab:osworld_res}, the resulting generalist model surpasses the performance of the individual specialist ensemble, demonstrating the effectiveness of a specialization-first strategy for achieving generalization. By learning from a broad range of software tasks, the generalist improves its reasoning and decision-making capabilities, acquiring transferable commonsense knowledge across domains.

% \noindent \textbf{Ablation Study of Specialist Training.} In Tab.\ref{tab:ablation}, we ablate the effectiveness of different components proposed using VScode success rate on OSWorld~\cite{xie2024osworld}. We first ablate the usage of World State Model to provide reward signals. The high precision in judge the success or failures of the actor agent's action compared to base model is crucial to the success self-evolution. Besides the reward signal, reinforcement-finetuning (RFT) also plays crucial roles compared to direct supervised fine-tuning (behavior cloning) as it encourage diversified reasoning with verifiable rewards to achieve more generalized task planning. After recognizing critical wrong action that lead to failure, Adversarial imitation enables the CUA to learn from failure, achieving additional improvements.

\begin{wraptable}{r}{0.6\linewidth} % r=右侧，l=左侧
% Table generated by Excel2LaTeX from sheet 'Sheet1'
% \begin{table}[h]
\vspace{-14pt}
\centering
\caption{Ablation of different configurations and their corresponding VScode success rates on OSWorld~\cite{xie2024osworld}. Using \judge as the reward model yields significant performance gains. We further compare different training strategies including supervised fine-tuning (behavior cloning), GRPO, and Adversarial Imitation (AI).}
\scalebox{0.66}{
\begin{tabular}{cccccc}
\toprule
\textbf{Qwen2.5VL-72B} & \textbf{\judge} & \textbf{SFT (BC)} & \textbf{GRPO} & \textbf{AI} & \textbf{VScode SR} \\
\midrule
 &  &  &  &  & 13.0 \\
\checkmark &  & \checkmark &  &  & 10.1 \\
\checkmark &  &  & \checkmark &  & 11.6 \\
 & \checkmark & \checkmark &  &  & 23.2 \\
 & \checkmark & \checkmark &  & \checkmark & 30.4 \\
 & \checkmark &  & \checkmark &  & 34.8 \\
 & \checkmark &  & \checkmark & \checkmark & 37.7 \\
\bottomrule
\end{tabular}
}
\vspace{-14pt}
\label{tab:ablation}
\end{wraptable}

\paragraph{Ablation Study of Specialist Training.}  
In Tab.~\ref{tab:ablation}, we present an ablation study on the effectiveness of various components in our training pipeline, using the success rate on VSCode from OSWorld~\cite{xie2024osworld} as the evaluation metric. 
First, we ablate the use of the World State Model for reward signal generation. Its high precision in judging the success or failure of the actor agent's actions—compared to using a base model—is shown to be essential for effective self-evolution. 
In addition to reward quality, reinforcement fine-tuning (RFT) also proves critical. Compared to direct supervised fine-tuning (behavior cloning), RFT encourages more diverse and exploratory reasoning patterns under verifiable rewards, enabling more generalized task planning. 
Finally, incorporating adversarial imitation to penalize critical failure-inducing actions allows the CUA to learn from its mistakes, yielding additional performance gains. This highlights the importance of learning not only from successful behaviors but also from failure signals.

% \noindent \textbf{Ablation Study of Generalist Training.}

% \begin{wrapfigure}{r}{0.5\linewidth} % r=右侧，l=左侧
%   \vspace{-10pt} % 根据需要微调
%   \includegraphics[width=\linewidth]{figures/process_curve.pdf}
%   \caption{\textbf{The Average Precision on AgentRewardBench~\cite{lu2025agentrewardbench}}, where GUI-Judge exhibits an improvement in AP as the number of input middle states increases, showing a similar trend to that of the closed sourced GPT-4o~\cite{hurst2024gpt} when compared with its base model.}
%   \label{fig:process_curve}
%   \vspace{-10pt} % 根据需要微调
% \end{wrapfigure}

% \section{Conclusion}

% In this work, we introduce \methodname, an autonomous computer use agent (CUA) exploration system that learns from its own experience on specific software. With a strong World State Model providing step level reward signal and curated corresponding RL frame work encouraging free form reasoning process while learning from trails and errors. The CUA successfully evolves into specialist on specific software. Further specialist-to-generalist training strategy helps us achieve a strong generalist on all the software. With Computer software being an ideal regularized virtual world, we believe this work can en-light future work in agentic system in game playing and embodied environments.

\section{Conclusion}

In this work, we introduce \methodname, an autonomous Computer Use Agent (CUA) exploration system that learns from its own experience on specific software. Powered by a robust World State Model that provides step-level reward signals, and a carefully designed reinforcement learning framework that encourages free-form reasoning through trial and error, the CUA is able to evolve into a specialist for individual software platforms. Furthermore, a specialist-to-generalist training strategy enables the development of a strong generalist agent capable of operating across multiple software environments. Given that computer software constitutes a highly regularized virtual world, we believe this work can inspire future research on agentic systems in both gaming and real world embodied environments.

\textbf{Limitations and future work.} While promising, our work still has several unresolved limitations. Firstly, our self evolving agent system is bounded by GUI-Judge to provide reliable reward signal instead of real signal from environment. As its still challenging to learning from sparse reward signal in complex environment. Secondly, though we tested on relatively complex and novel software like libreoffice-tools and GIMP. The task is still relatively simple as it only takes a human expert less than 20 step to accomplish. How to adapt the system to achieve hours-long workflow in even more challenging software used by real human expert are thus interesting future directions.

\clearpage
\bibliography{egbib}
\bibliographystyle{plain}

\clearpage
\appendix
% \section{Details of World State Model.}
% \label{sup:gui_judge_train}
% Provide CUAs with reliable reward signal is crucial for the agent system (actor and critic). As illustrated in Fig.\ref{fig:process_curve} and Sec.\ref{sec:gui_judge}, it is important to fine-tune a reward model take all the history states (screenshots) as input and. To close the large gap with commercial model like GPT-4o~\cite{hurst2024gpt} and incentivize local model to provide judgement based on step by step detailed analysis from high-res screen shot. We use GPT-4o~\cite{gpt4} to generate step level judgment using prompt in Fig.\ref{fig:os_prompt} on 860 trajectories generated by two different agents UI-TARS~\cite{qin2025uitars} and Gemini-2.5-pro~\cite{team2023gemini} on 43 feasible episode of Chrome from OSWorld~\cite{agashe2024agent}. This is cooperated with detailed  In purpose of incentivizing reward model for sequencial checking on history screenshot with high resolution (15+ 1980x1080), we use Qwen2.5-VL-7B as strong base model, finetuning is done use LoRA with rank=128 with freezed vision encoder to mitigate overfitting. Training is done on 8 NVIDIA-A100 80 GPUs for 2000 iteration with learning rate set to 2e-4. 
\section{World State Model}
\label{sup:gui_judge}
The World State Model (WSM) is a central component of SEAgent, responsible for understanding visual state changes and evaluating the effectiveness of the agent's actions.

\subsection{Model Architecture and Operation}
The WSM is built upon the Qwen2.5-VL-7B vision-language model. It operates in two distinct modes, each with a specific input-output structure to perform different tasks:
\begin{enumerate}
    \item \textbf{Trajectory Judgment:} 
    \begin{itemize}
        \item[\textbf{Input:}] A sequence of screenshot images captured during an episode.
        \item[\textbf{Output:}] Short captions for each screenshot, the reasoning process for the judgment, and a structured judgment dictionary (containing fields such as \texttt{Correctness}, \texttt{Redundant}, and \texttt{First Error Step}, as detailed in Figure~7 of the supplementary material).
    \end{itemize}
    \item \textbf{State Change Description:} 
    \begin{itemize}
        \item[\textbf{Input:}] Two screenshot images, one from before and one after a single action was executed.
        \item[\textbf{Output:}] A detailed description of the visual differences between the two images.
    \end{itemize}
\end{enumerate}

\subsection{Fine-Tuning Dataset and Process}
To equip the WSM with these capabilities, a specialized dataset was constructed for fine-tuning.
\paragraph{Data Construction} The data construction process is as follows:
\begin{enumerate}
    \item \textbf{Trajectory Sampling:} A Computer Using Agent (CUA), powered by UI-TARS and Gemini-2.5-Pro, was used to sample trajectories from 43 feasible tasks in Google Chrome within the OSWorld benchmark. These trajectories were saved as screenshot sequences.
    \item \textbf{GPT-4o Annotation:} Using the prompts detailed in Figures 6 and 7 of the supplementary material, GPT-4o was employed to annotate the sampled trajectories, generating judgments and screenshot captions. Only samples where the judgment matched the ground truth from OSWorld evaluation protocols were retained, resulting in 860 high-quality annotated trajectories.
    \item \textbf{Change Description Data:} An additional 1,000 pairs of (before action, after action) screenshots were sampled. GPT-4o was used to generate detailed descriptions of the differences, creating a 1,000-sample Change Description (CD) dataset.
\end{enumerate}

\paragraph{Fine-Tuning Process}
\label{sup:gui_judge_train}
The fine-tuning was performed using the Llama-Factory framework on 8 NVIDIA A100 (80G) GPUs for 2,000 iterations. A learning rate of $2 \times 10^{-5}$ was used, and LoRA (rank=128) was employed for parameter-efficient fine-tuning.
The 860 annotated trajectories serve as the core training data for teaching the model trajectory judgment, captioning, and reasoning. The 1,000-sample CD dataset acts as auxiliary data, specifically to encourage the model to focus on fine-grained visual differences, which enhances its overall state understanding. As shown in Table 1 of the main paper, incorporating CD data significantly boosts judgment performance. The two datasets were combined for training without any special re-weighting.

\subsection{Reward Generation from Trajectory Analysis}
The trajectory judgment capability of the WSM is the core source of the reward signal for reinforcement learning. After an agent executes a full trajectory $\mathcal{H} = \{s_0, a_0, s_1, a_1, \dots, s_{\text{final}}\}$, the WSM analyzes it and outputs a structured judgment. Based on this output, actions within the trajectory are dynamically labeled as either positive actions ($a_T$) or failure actions ($a_F$):
\begin{itemize}
    \item \textbf{Fully Successful Trajectory:} If \texttt{Correctness} is `True` and there are no \texttt{Redundant} steps, all actions $a$ in the trajectory are labeled as $a_T$.
    \item \textbf{Successful but Inefficient Trajectory:} If \texttt{Correctness} is `True` but \texttt{Redundant} steps begin at step $k$, all actions prior to step $k$ are labeled as $a_T$.
    \item \textbf{Failed Trajectory:} If \texttt{Correctness} is `False` and the \texttt{First Error Step} is $e$, all actions prior to step $e$ are labeled as $a_T$, while the erroneous action $a_e$ is labeled as $a_F$.
\end{itemize}
These dynamically labeled $a_T$ and $a_F$ actions constitute the reward signals for the RL pipeline. During training, the actor predicts an action $a_t$ based on the history $\{a_0, s_0, \dots, s_t\}$ and uses these labels to calculate rewards.

\section{Curriculum Generator}
The Curriculum Generator is designed to dynamically produce tasks of increasing difficulty and diversity, guiding the agent through a systematic exploration of the software's capabilities.

\subsection{Task Generation Mechanism}
The workflow of the Curriculum Generator is detailed in the pseudocode in our supplementary material. Its core idea is to leverage the WSM's analysis of completed tasks to generate new ones. The process, illustrated by the "add a rectangle" example from Figure 5, involves three main steps:
\begin{enumerate}
    \item \textbf{Analysis and Feedback:} The agent successfully completes an initial task, "add a rectangle." The WSM analyzes the execution trajectory and extracts two key pieces of information: a task evaluation (\texttt{Exam}) and a list of observed state changes (\texttt{CD\_list}).
    \begin{quote}
        \texttt{CD\_list}: \{"add a rectangle": ["The Edit bar is expanded...", "The cursor has changed into a cross...", "A blue box appears on the screen with side bars showing properties such as fill, line, color, width, transparency, and corner style..."], ...\} \\
        \texttt{Exam}: [\{"task": "add a rectangle", "status": "success"\}, ...]
    \end{quote}
    \item \textbf{Knowledge Integration and Task Generation:} The \texttt{CD\_list} and \texttt{Exam} are fed into the Curriculum Generator. It distills new knowledge, such as "properties of a rectangle," and integrates it into its internal \texttt{Software guidebook}. Based on this new knowledge, it generates more challenging tasks like "Add a green rectangle" or "Add a red rectangle with 50\% transparency," which are then added to the task buffer.
    \item \textbf{Iterative Learning:} In the next RL phase, the agent samples from this updated, more challenging task buffer. The continuously enriched \texttt{Software guidebook} acts as the system's long-term memory, driving the Curriculum Generator to propose increasingly sophisticated and unexplored tasks in subsequent rounds, thereby guiding the agent toward mastery.
\end{enumerate}

\section{Details of Curriculum Generator.}
\subsection{Exemplar Case during Task Evolution.}
\label{sup:data_pipe}
\begin{figure*}[h]
  \centering
  \includegraphics[width=1.0\linewidth]{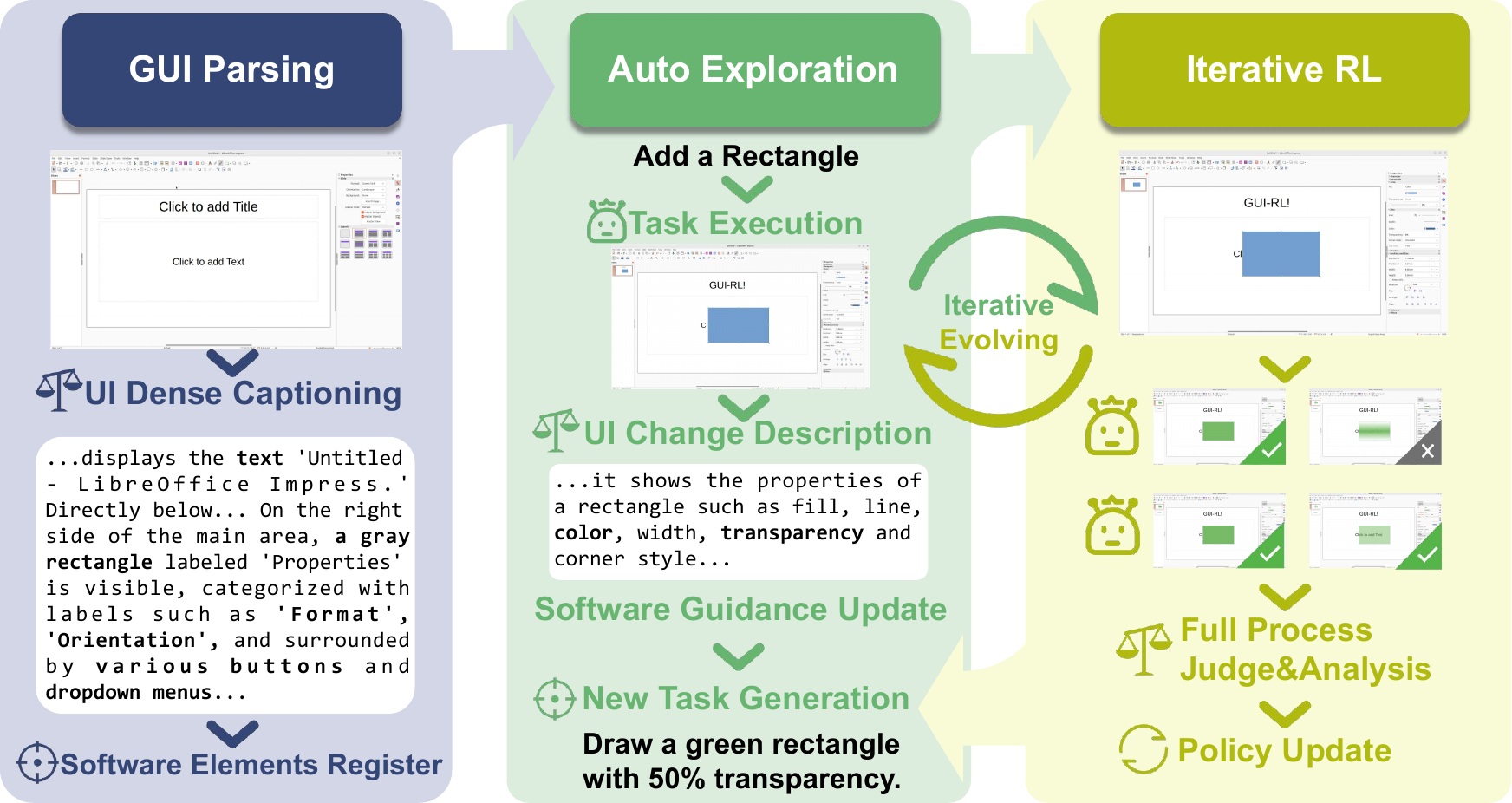}
   \caption{\textbf{\methodname autonomous exploration pipeline.} The agent (policy model) and World State Model iteratively generate new task and perform RL to become a specialist in novel software.}
   \label{fig:data_pipeline}
\end{figure*}

We provide an exemplar case of our task evolution pipeline in Fig.~\ref{fig:data_pipeline}, demonstrated using LibreOffice Impress. Initially, the World State Model parses a screenshot of the Impress interface into detailed captions describing the layout and individual buttons. The Task Generator then produces an initial task set, $\mathcal{I}_0 = \{I_0^{(1)}, I_0^{(2)}, \ldots\}$, and summarizes the initial software guidance memory $U_0$. The initial agent executes tasks in $\mathcal{I}_0$, such as ``Add a Rectangle,'' while the World State Model evaluates these actions, providing judgments and detailed descriptions of resulting changes. As shown in the Auto-Exploration stage, this includes generating captions for newly appeared property panels and assessing execution success. The Task Generator incorporates feedback on execution success and newly revealed properties (e.g., transparency) to evolve new tasks, such as ``Draw a green rectangle with 50\% transparency.'' This process iteratively improves through reinforcement learning, enabling continuous task evolution and agent self-improvement.

\subsection{Comparative Analysis of Instruction Generation Strategies.}
\label{sup:comparision_task_gen}
To validate the effectiveness of our Curriculum Generator, we conducted a comparative analysis against state-of-the-art instruction generation methods, namely those from NNetNav~\cite{murty2025nnetnavunsupervisedlearningbrowser} and WebRL~\cite{qi2024webrl}.

\paragraph{Experimental Setup}
We adapted the official code and prompts from these prior works from web environments to general software applications. To ensure a fair comparison of the curriculum quality, for each strategy, we employed two leading LLMs: the open-source Qwen2.5-72B~\cite{bai2025qwen2} and the proprietary Gemini-2.5-Pro~\cite{google2025gemini25preview}. The tasks generated by each strategy were used to train an RL agent (using GRPO only), with reward signals uniformly provided by our fine-tuned WSM. The evaluation was performed on two applications: VSCode from OSWorld (a standard software) and Celestia from ScienceBoard~\cite{sun2025scienceboardevaluatingmultimodalautonomous} (a more challenging, out-of-domain scientific application). The primary metric was the task success rate.

\paragraph{Results and Discussion}
The results are presented in Table~\ref{tab:comparison_en}.
\begin{table}[h!]
\centering
\caption{Success rate (\%) comparison of different task generation strategies on two software applications.}
\label{tab:comparison_en}
\begin{tabular}{rccc}
\toprule
\textbf{Task Generation Strategy} & \textbf{LLM}             & \textbf{VSCode} & \textbf{Celestia} \\ \midrule
WebRL                         & Qwen2.5-72B              & 27.5            & 0.00              \\ 
WebRL                         & Gemini2.5-Pro-thinking   & 36.2            & 3.03              \\ 
NNetNav                       & Qwen2.5-72B              & 34.6            & 0.00              \\ 
NNetNav                       & Gemini2.5-Pro-thinking   & 43.6            & 5.05              \\ 
Curriculum Generator (Ours)   & Qwen2.5-72B              & 37.7            & 9.09              \\ 
Curriculum Generator (Ours)   & Gemini2.5-Pro-thinking   & 42.3            & 12.12             \\ 
\bottomrule
\end{tabular}
\end{table}

As shown, the reverse instruction generation strategy from NNetNav~\cite{murty2025nnetnavunsupervisedlearningbrowser} is highly effective on the in-domain application (VSCode), demonstrating high data generation efficiency by producing successful trajectories. However, a critical trade-off was observed: this approach tends to generate many similar tasks, limiting its ability to explore the full breadth of the software's functionalities. This limitation becomes more pronounced when the task generator is unfamiliar with the target software, as seen in the OOD Celestia environment.

In contrast, our guidebook-based method, while having a lower initial data generation efficiency, excels at systematic exploration. It builds structured knowledge of the software from scratch, making it more robust for tackling novel applications. This is evidenced by its superior performance on the more challenging Celestia software.

We conclude that these two strategies are complementary. Reverse instruction generation can efficiently exploit known functionalities, while our guidebook-based method can systematically explore new ones and help the task generator build a more comprehensive understanding of the target software. A hybrid approach combining both strategies is a promising direction for future work.

% We provide a exemplar case of our task evolution pipeline in Fig.\ref{fig:data_pipeline} in Libreoffice-Impress software. At the beginning, the World State Model will parse the initial screen shot of Impress into detailed captions of layout and each button. The Task Generator will generate the initial task set $\mathcal{I_0} = \{I_0^{(1)}, I_0^{(2)}, ...\}$ and summaries the initial software guidance memory $U_0$. The initial agent while take actions to execute $I_0$, for example "Add a Rectangle." with World State Model provide judges and detailed change descriptions as shown in Auto Exploration stage, caption on new emerged panel of the properties of the rectangle, with judgment of success of execution. The Task Generator will takes the success of execution and newly emerged properties of the rectangle (like transparency) into account and evolve the new task of "Draw a green rectangle with 50\% transparency". These process with RL wile perform iteratively to achieve self evolving improvements.

% Section on TARS-1.5 Test
\section{Test on TARS-1.5}
Our work focuses on enabling agents to adapt to out-of-domain (OOD) and novel software where human-labeled data is not available. To test this, we applied our SEAgent pipeline to the UI-TARS-1.5~\cite{qin2025uitars} model on two distinct benchmarks. On OSWorld~\cite{xie2024osworld}, we observed moderate performance gains. We hypothesize this is because UI-TARS-1.5's training data already targeted OSWorld, making it a familiar, in-domain environment for the base model. However, on the ScienceBoard \cite{sun2025scienceboard} benchmark---a suite of scientific applications that are truly novel to UI-TARS-1.5---our pipeline delivers significant and substantial improvements. This strongly validates our core claim: SEAgent is most impactful when performing self-evolution learning on truly OOD software. We excluded two of the six ScienceBoard applications---Lean and TeX---as they are primarily text- and code-based software for mathematics and typesetting, which are not suitable for evaluating a GUI-centric agent like UI-TARS.

% Note: Please replace 'scienceboard_ref' with your actual bibliography key.
% For the table, we recommend using the booktabs package for professional-quality tables.
\begin{table}[h!]
\centering
\caption{Performance comparison on OSWorld and ScienceBoard benchmarks. Scores represent success rates (\%).}
\label{tab:tars_performance}
\scalebox{0.8}{
\begin{tabular}{lccccccr}
\toprule
& \multicolumn{3}{c}{\textbf{OSWorld}} & \multicolumn{4}{c}{\textbf{ScienceBoard}} \\
\cmidrule(lr){2-4} \cmidrule(lr){5-8}
\textbf{Model} & \begin{tabular}[c]{@{}c@{}}LibreOffice\\Impress\end{tabular} & \begin{tabular}[c]{@{}c@{}}LibreOffice\\Writer\end{tabular} & GIMP & ChamerX & GrassGIS & KAlgebra & Celestia \\
\midrule
UI-TARS-1.5-7B-DPO & 19.15 & 33.04 & 51.54 & 12.41 & 0.00 & 11.61 & 4.85 \\
UI-TARS-1.5-7B-DPO+SEAgent & 23.83 & 35.65 & 56.92 & 23.45 & 10.59 & 21.29 & 11.52 \\
\bottomrule
\end{tabular}
}
\end{table}

% Section on Sensitivity Analysis
\section{Sensitivity Analysis on Key Hyperparameters}
We conducted a sensitivity analysis on key hyperparameters to evaluate their impact on the SEAgent pipeline. For model sampling, we set the temperature $t=0$ for better reproducibility. We analyze two specific parameters: the number of generated tasks and the number of change descriptions. The results are presented in Table~\ref{tab:sensitivity_analysis} and discussed below.

\begin{table}[h!]
\centering
\caption{Sensitivity analysis for key hyperparameters in the SEAgent pipeline, evaluated on VSCode. The metric is Success Rate (\%).}
\label{tab:sensitivity_analysis}
\scalebox{1.0}{
\begin{tabular}{cc|cc}
\toprule
\textbf{\# Tasks Generated} & \textbf{VScode SR} & \textbf{\# Change Descriptions} & \textbf{VScode SR} \\
\midrule
30  & 31.88 & 30  & 33.33 \\
50  & 36.23 & 50  & 37.68 \\
100 & 37.68 & 100 & 37.68 \\
200 & 37.68 & 200 & 34.78 \\
\bottomrule
\end{tabular}
}
\end{table}

\paragraph{Number of Generated Tasks} This parameter controls the breadth of exploration in each learning cycle. As shown in our analysis, performance improves as more diverse tasks are generated, eventually plateauing around 100 tasks.

\paragraph{Number of Change Descriptions} This parameter controls how much new information the generator receives to update its "software guidebook." We found a clear trade-off: A sufficient number of descriptions (50--100) is essential for the generator to learn about new UI functionalities and create meaningful, unexplored tasks. However, providing too many descriptions (e.g., 200) creates an overly long context for the LLM, which degrades the quality of task generation and hurts final performance.

\section{Ablation on the Loss Balance Factor.}
\label{sup:loss_factor}
In Sec.\ref{sec:rl}, we use $\gamma$ to balance the ratio of two loss item: adversarial imitation that learn from error and GRPO that learn to achieve success. We ablate the choice of $\gamma$ in Tab.\ref{tab:beta_choice}, according to which we set $\gamma = 0.2$ in main experiments.

\begin{table}[h]
\centering
\begin{tabular}{c|cccccc}
\toprule
\textbf{$\gamma$}         & 0.0   & 0.1 & 0.2 & 0.3 & 0.5 & 0.8 \\
\midrule
Success Rate (\%) & 34.8 & 36.2 & 37.7 & 31.9 & 26.1 & 23.1 \\
\bottomrule
\end{tabular}
\caption{VScode Success Rate on OSWorld~\cite{xie2024osworld} under different loss balance factor $\gamma$ values.}
\label{tab:beta_choice}
\end{table}

\section{Reward Function for Different Actions.}

\begin{table}[h]
\centering
\scalebox{0.57}{
\begin{tabular}{l|l|l}
\toprule
\textbf{Action Type} & \textbf{Description} & \textbf{Distance-based Reward} \\
\midrule
\texttt{click}, \texttt{left\_single}, \texttt{right\_single}, \texttt{hover} 
    & Click or hover on a location 
    & Normalized L1 distance between predicted and ground-truth coordinates \\
\texttt{left\_double}, \texttt{double\_click} 
    & Double click on a region 
    & Normalized L1 distance between clicked coordinates \\
\texttt{drag}, \texttt{select} 
    & Drag from start box to end box 
    & Intersection over Union (IoU) between predicted and ground-truth boxes \\
\texttt{type} 
    & Type textual input 
    & Character-level BLEU score between predicted and ground-truth text \\
\texttt{hotkey} 
    & Press multiple keys at once 
    & Character-level BLEU score between predicted and ground-truth key combinations \\
\texttt{press} 
    & Press a single key 
    & Character-level BLEU score between predicted and ground-truth key \\
\texttt{scroll} 
    & Scroll in a certain direction 
    & Character-level BLEU score between predicted and ground-truth direction \\
\texttt{move\_mouse} 
    & Move mouse to a specific location 
    & Normalized L1 distance between predicted and ground-truth coordinates \\
\texttt{highlight} 
    & Highlight a rectangular UI region 
    & IoU between predicted and ground-truth region \\
\texttt{copy}, \texttt{paste} 
    & Clipboard operations 
    & BLEU score between copied/pasted content \\
\texttt{wait} 
    & Explicit wait command 
    & Fixed reward + 1 \\
\texttt{finished}, \texttt{finish\_task} 
    & Finish current task/trajectory 
    & Fixed reward + 1 \\
\bottomrule
\end{tabular}
}
\caption{Reward computation for each action type in GUI agent}
\label{tab:reward_list}
\end{table}

\section{Data Statistics during Iterative Reinforcement Learning.}
\label{sup:data_stats}

\begin{table}[h]
\centering
\begin{tabular}{lcccc}
\toprule
        & Phase0 & Phase1 & Phase2 & Phase3 \\
\midrule
VSCode  & 112/39 & 282/83 & 161/34 & 98/55  \\
GIMP    & 104/51 & 309/90 & 183/50 & 95/52  \\
Impress & 102/44 & 290/92 & 185/61 & 87/51  \\
VLC     &  85/29 & 114/41 & 160/48 & 53/27  \\
Writer  & 123/62 & 278/101 & 201/69 & 101/43 \\
\bottomrule
\end{tabular}
\caption{Number of episode (Success/Failure) across four phases for different software tools during self-evolution. Each episode contains 8.8 multi-turn conversions in average.}
\label{tab:phase_data}
\end{table}

\section{Detailed Prompt Templates.}
\begin{figure*}[h]
  \centering
  \includegraphics[width=1.0\linewidth]{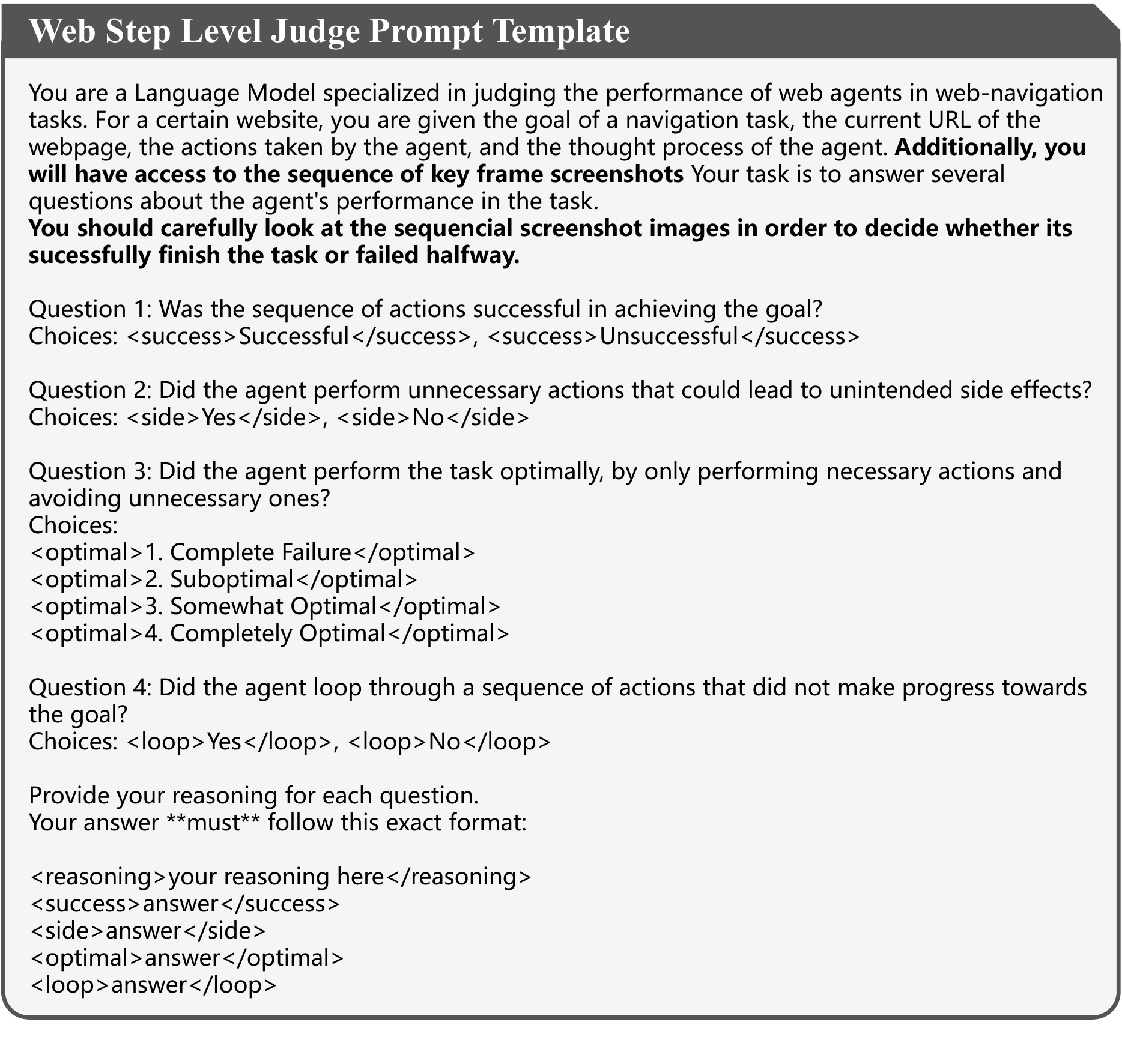}
   \caption{\textbf{Prompt Template of GUI-Judge for web agent trajectories evaluations} with history screenshots as input, its difference with default prompt of AgentRewardBench~\cite{lu2025agentrewardbench} is highlighted in bold.}
   \label{fig:web_prompt}
\end{figure*}

For evaluation on AgentRewardBench~\cite{lu2025agentrewardbench}, we use their official template for final state screenshot only testing and modified prompt in Fig.\ref{fig:web_prompt} for entire process (or sampled middle screenshots) testing. 

\begin{figure*}[h]
  \centering
  \includegraphics[width=1.0\linewidth]{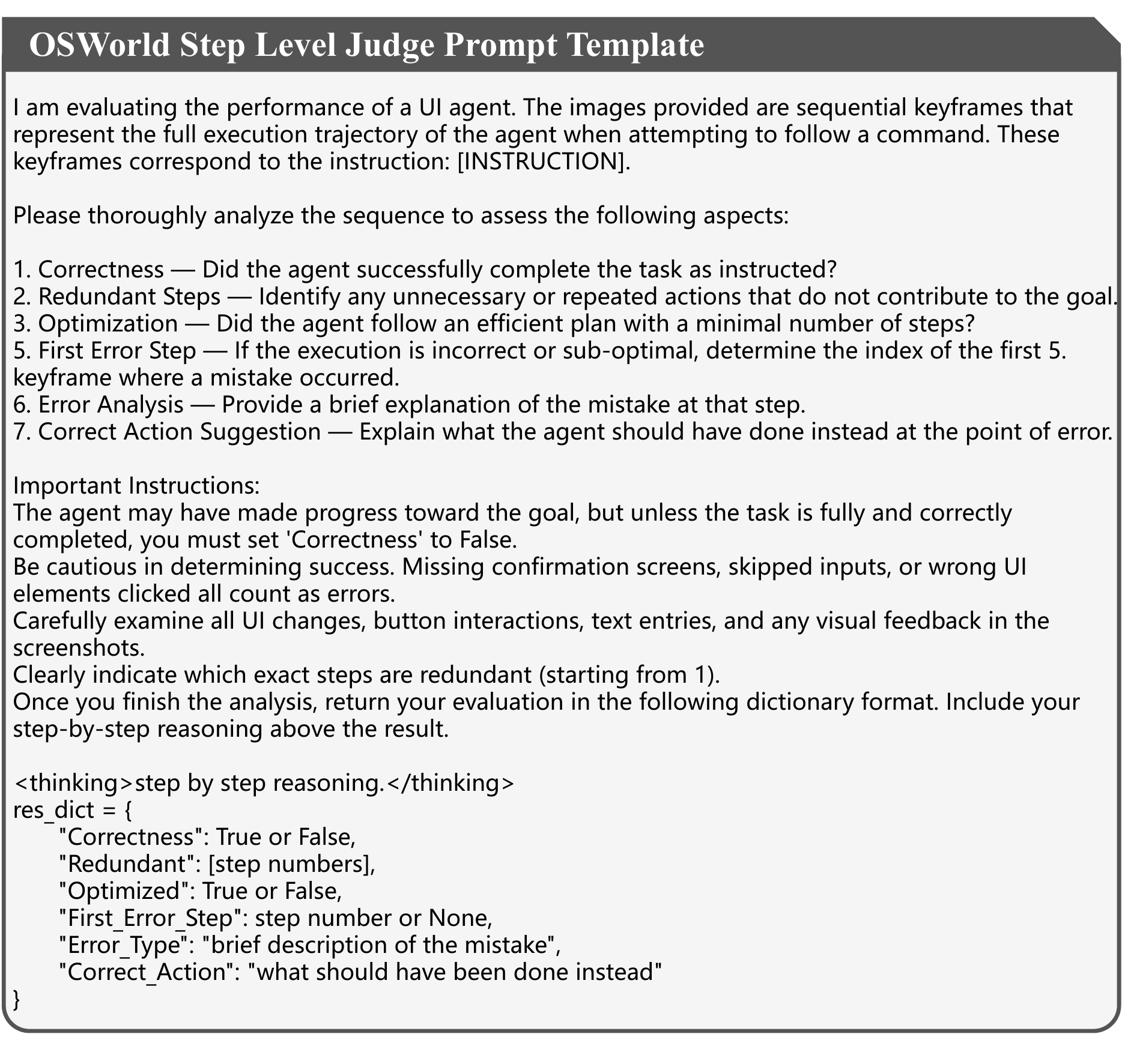}
   \caption{\textbf{Prompt Template of GUI-Judge for OSWorld~\cite{xie2024osworld} trajectories}, which prompts judge model to provide step level reward signal.}
   \label{fig:os_prompt}
\end{figure*}

For evaluation on OSWorld Sampled trajectories, we use prompt in Fig.\ref{fig:os_prompt} to prompt GPT-4o to provide step level judges, the sampled judges on Chrome in OSWorld~\cite{xie2024osworld} serves as training data of GUI-Judge. This template is also used in training GUI-Judge and at inference time in autonomous exploration stage.

\begin{figure*}[h]
  \centering
  \includegraphics[width=1.0\linewidth]{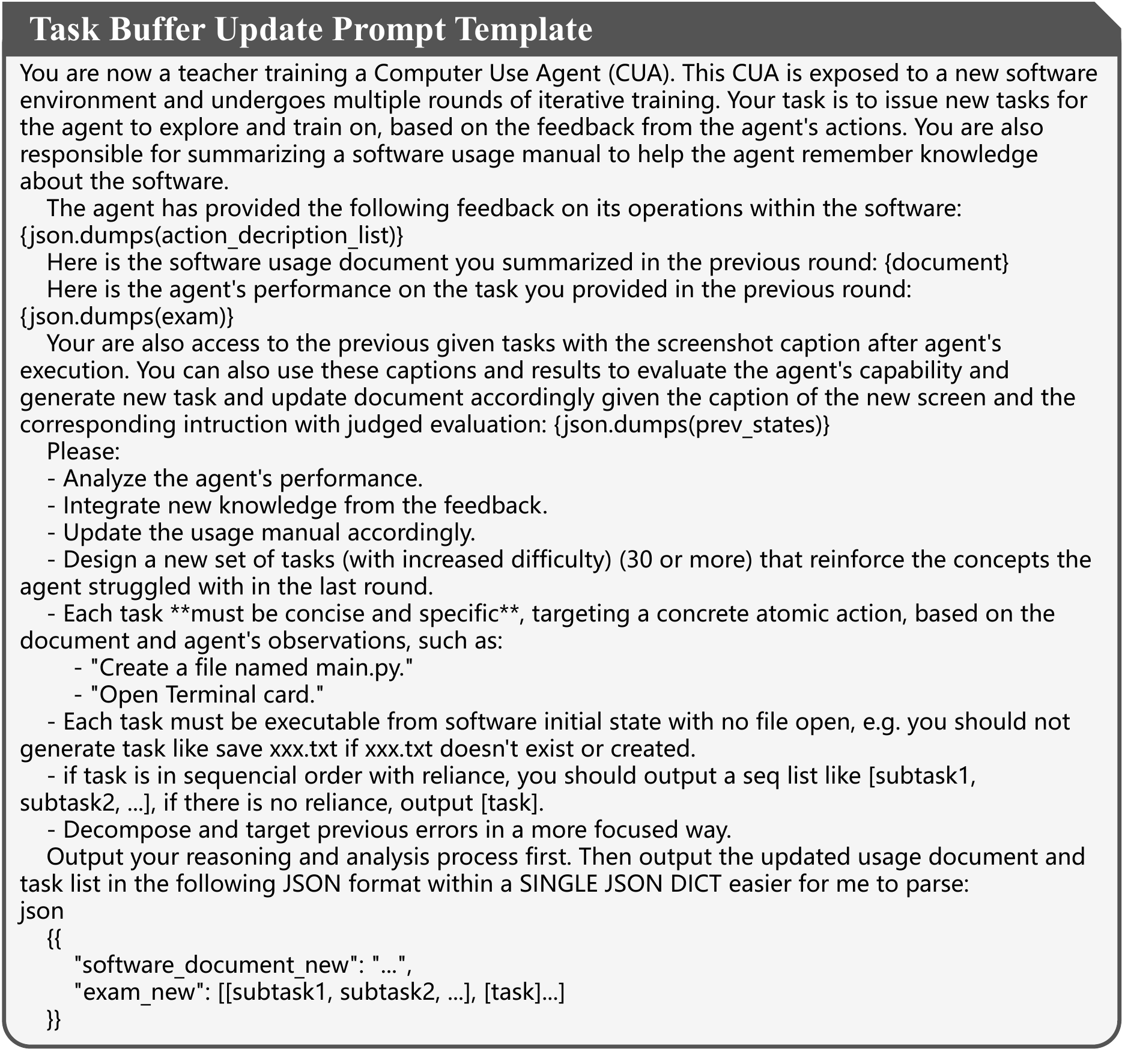}
   \caption{\textbf{Prompt Template for task buffer update}, which generates new tasks in a curriculum manner and update software documents. The new tasks are used for actor to perform next phase of RL.}
   \label{fig:task_buffer_prompt}
\end{figure*}

For navigator, we use prompt template in Fig.\ref{fig:task_buffer_prompt}, which takes previous software usage manual and the performance of actor agent evaluated by judge (Empty if in initial phase.) as well as detailed exploration caption as input and output the updated usage manual as well as new task for agent to execute.

\section{Self documented usage manual on different software during exploration.}
\label{sup:usage_manual}
\begin{figure*}[h]
  \centering
  \includegraphics[width=1.0\linewidth]{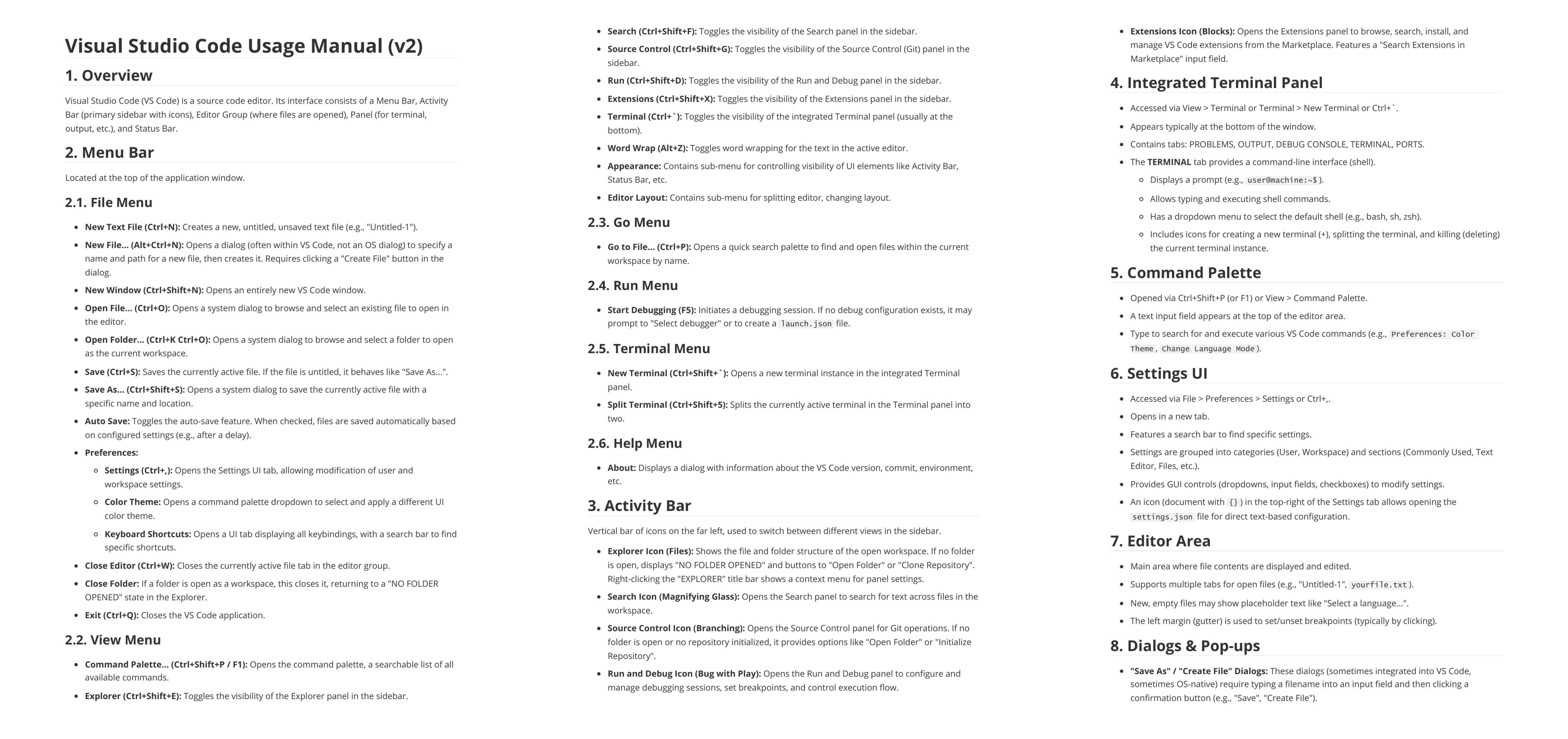}
   \caption{\textbf{Automatically generated usage manual during self exploration} on VScode.}
   \label{fig:vscode_doc}
\end{figure*}

\begin{figure*}[h]
  \centering
  \includegraphics[width=1.0\linewidth]{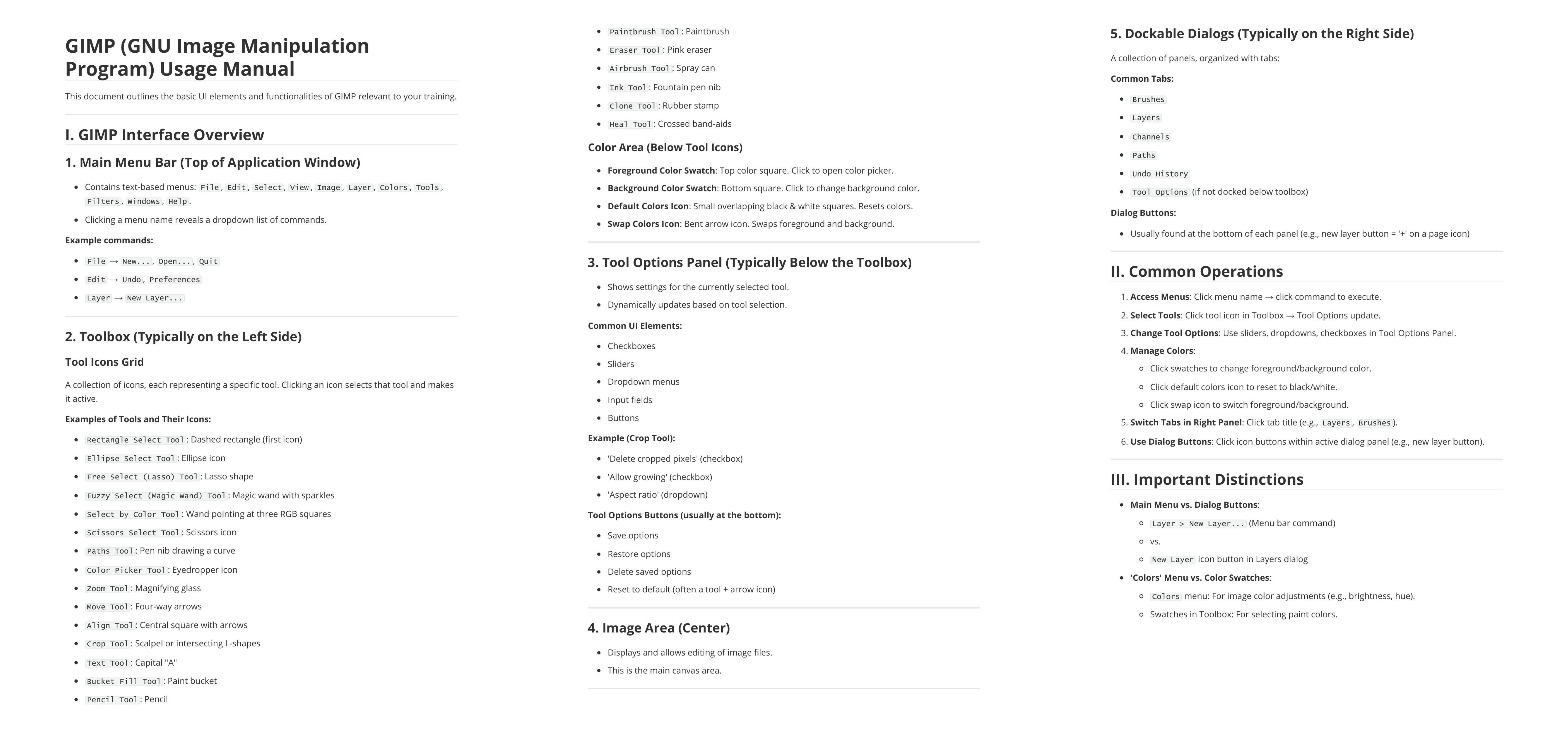}
   \caption{\textbf{Automatically generated usage manual during self exploration} on GIMP.}
   \label{fig:gimp_doc}
\end{figure*}

\begin{figure*}[h]
  \centering
  \includegraphics[width=1.0\linewidth]{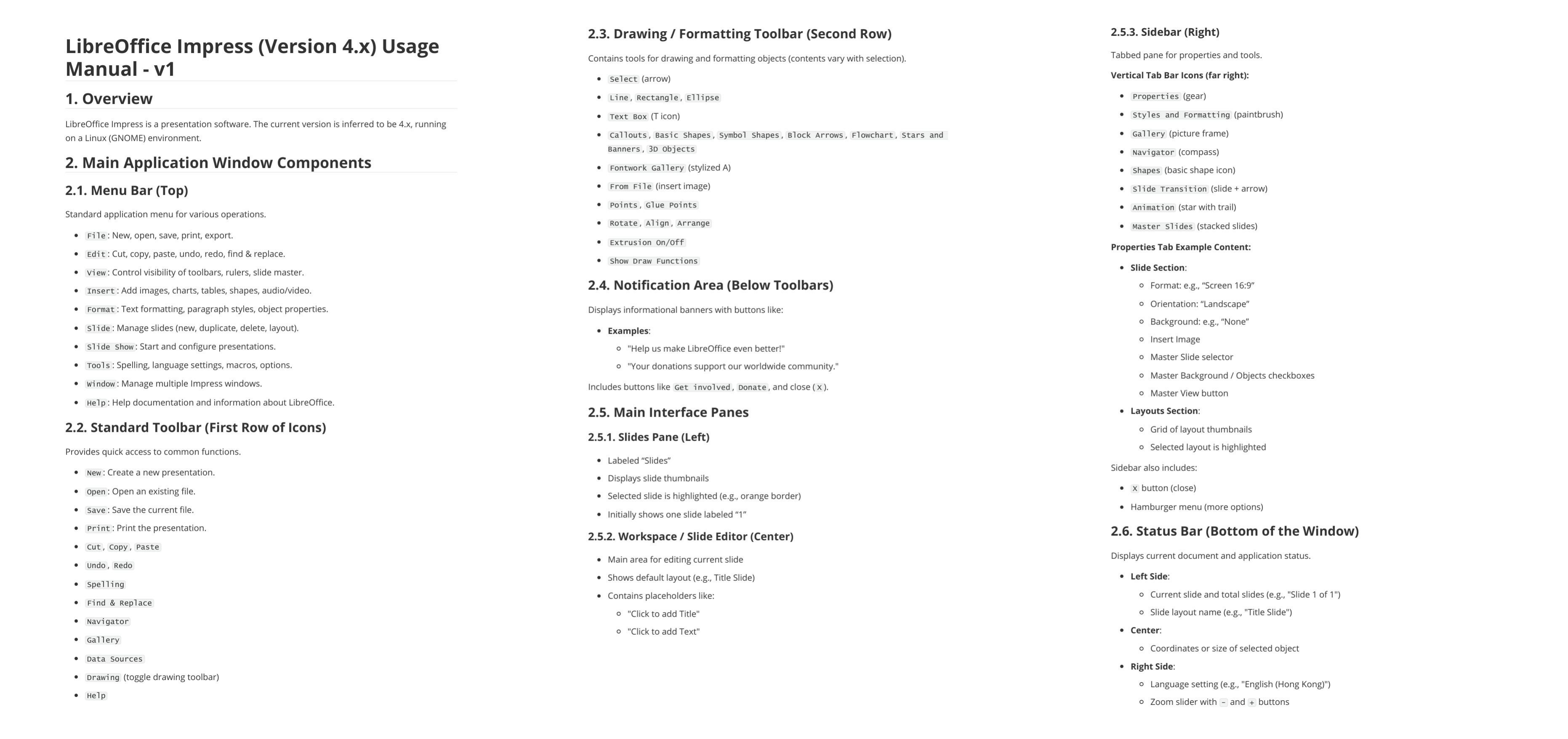}
   \caption{\textbf{Automatically generated usage manual during self exploration} on LibreOffice\_Impress.}
   \label{fig:impress_doc}
\end{figure*}

\begin{figure*}[h]
  \centering
  \includegraphics[width=1.0\linewidth]{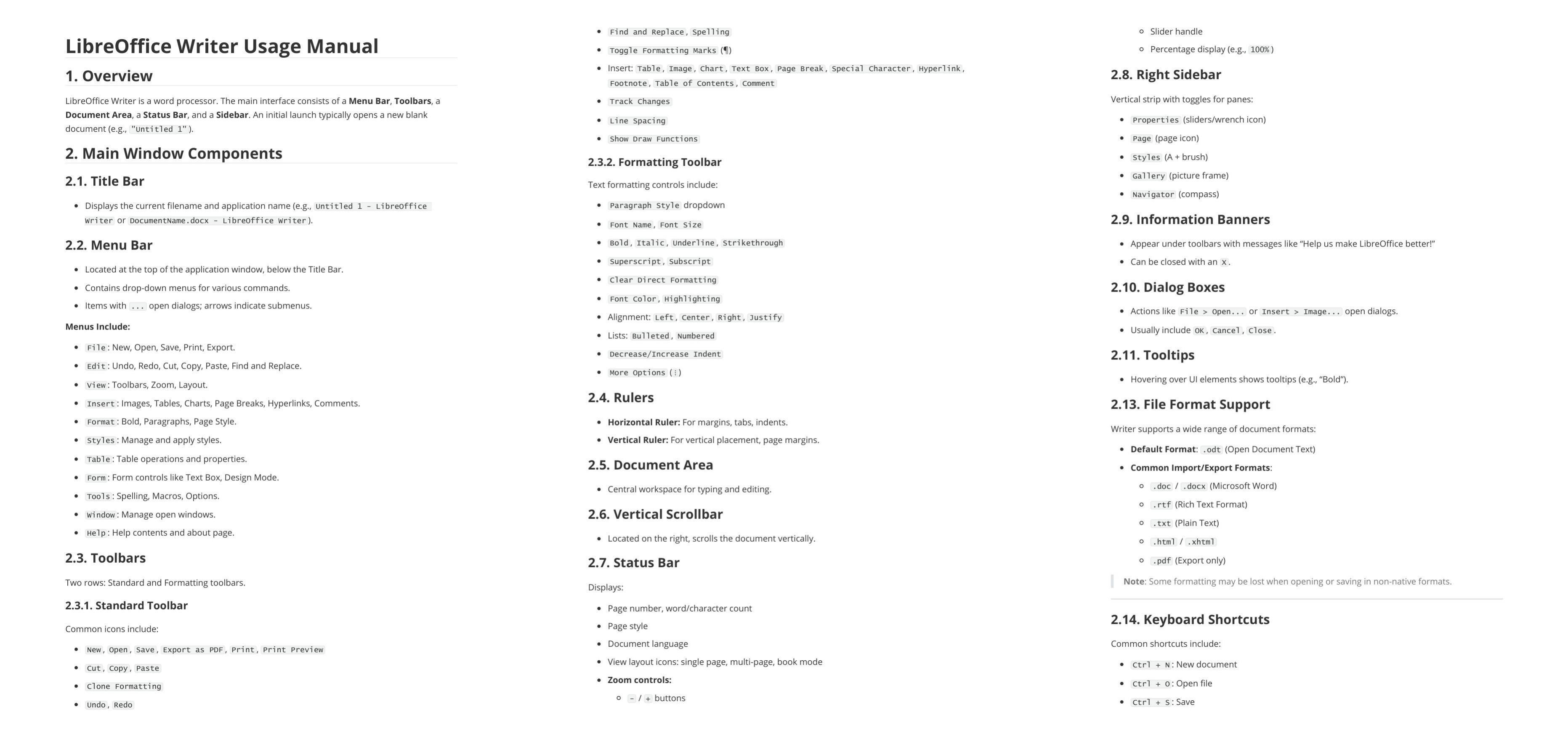}
   \caption{\textbf{Automatically generated usage manual during self exploration} on LibreOffice\_Writer.}
   \label{fig:writer_doc}
\end{figure*}

In Fig.\ref{fig:vscode_doc} Fig.\ref{fig:impress_doc}, Fig.\ref{fig:gimp_doc}, Fig.\ref{fig:writer_doc}, we demonstrate the self-documented usage manuals of the navigator (Qwen2.5-72B~\cite{yang2024qwen2}) in the exploration and learning system introduced in Sec.\ref{sec:autonomous_exploration_pipeline}.

\section{Broader Impacts}
\label{sup:broader_impacts}
\textbf{Potential positive societal impacts:}
SEAgent introduces a self-evolving paradigm for Computer Use Agents (CUAs), enabling them to autonomously learn and adapt to previously unseen software without human supervision. This significantly reduces the need for extensive manual data annotation and domain-specific customization, allowing intelligent agents to assist users across a wide range of applications—including productivity tools, multimedia editing, and educational software. By automating repetitive tasks and providing guidance in complex software environments, SEAgent holds promise for improving accessibility, enhancing digital literacy, and reducing cognitive workload in both professional and everyday settings.

\textbf{Potential negative societal impacts:}
The capability of SEAgent to autonomously explore and operate complex software also introduces risks of misuse. Malicious actors might repurpose SEAgent for unauthorized software automation, such as automating account creation, spamming interfaces, or conducting surveillance via GUI interactions. In addition, as the agent learns from its own experience, there exists a risk that the agent may inadvertently inherit or amplify software-specific biases, potentially leading to unfair or inappropriate behaviors in sensitive applications (e.g., finance, legal automation). Mitigation strategies include controlled release of models, behavior filters during deployment, and incorporating safeguards in the World State Model to detect and prevent unintended or adversarial behavior.

\begin{algorithm}
\caption{SEAgent Specialized Self-Evolution Training Loop}
\label{alg:self-evolution}
\begin{algorithmic}[1]
\State \textbf{Input:} Initial policy $\pi_0$, World State Model $\mathcal{M}_{\text{state}}$, Curriculum Generator $\mathcal{M}_{\text{task}}$, Initial GUI state $S_0$

\State \textbf{1. Task Initialization}
\State $\mathcal{C}_0 \gets \text{CaptionGUI}(S_0)$ \Comment{Parse initial GUI layout (menu bar, buttons, etc.)}
\State $\mathcal{I}_0, U_0 \gets \mathcal{M}_{\text{task}}(\emptyset, \emptyset, \emptyset, \mathcal{C}_0)$ \Comment{Generate basic initial tasks and usage guide}

\noindent\textcolor{gray}{\rule{\linewidth}{0.3pt}}  % ← 横线分隔

\For{$p = 0$ to $P-1$} \Comment{2. Self-Evolution Phase Loop}

    \State \textbf{2.1 Autonomous Exploration}
    \State $\mathcal{D}_{\text{traj}} \gets \emptyset$
    \ForAll{$I \in \mathcal{I}_p$}
        \State $\tau \gets \text{ExecuteInstruction}(\pi_p, I)$ \Comment{Actor executes task in the virtual environment}

        \State \textbf{2.2 Effect Evaluation}
        \State $\mathcal{J}_I, \mathcal{C}_I \gets \mathcal{M}_{\text{state}}(\tau)$ \Comment{Step-level trajectory judgment and new state captions}
        \State $\mathcal{D}_{\text{traj}} \gets \mathcal{D}_{\text{traj}} \cup \{(\tau, \mathcal{J}_I, \mathcal{C}_I)\}$ \Comment{$\mathcal{J}_I$: a sequence of per-step feedback labels ($a_T$ or $a_F$)}
    \EndFor

    \noindent\textcolor{gray}{\rule{\linewidth}{0.3pt}}

    \State \textbf{2.3 Policy Update (RFT)}
    \State Split $\mathcal{D}_{\text{traj}}$ into:
    \State \quad $\mathcal{D}_\text{pos}$: steps labeled as positive $a_T$
    \State \quad $\mathcal{D}_\text{neg}$: steps labeled as negative $a_F$

    \State Compute GRPO loss on $\mathcal{D}_\text{pos}$:
    \State \quad $r(a, a_T) = \mathbb{I}[\text{type}(a)=\text{type}(a_T)] + r_{\text{dist}}(a, a_T)$
    
    \State Compute Adversarial Imitation loss on $\mathcal{D}_\text{neg}$:
    \State \quad $\mathcal{L}_{\text{AI}} = - \log \frac{\pi_\theta(a \mid s, I)}{\pi_{\text{ref}}(a_F \mid s, I)}$

    \State Total loss: $\mathcal{L}_{\text{total}} = \mathcal{L}_{\text{GRPO}} + \gamma \mathcal{L}_{\text{AI}}$
    \State $\pi_{p+1} \gets \text{Update}(\pi_p, \mathcal{L}_{\text{total}})$

    \noindent\textcolor{gray}{\rule{\linewidth}{0.3pt}}

    \State \textbf{2.4 Task Update}
    \State $\mathcal{I}_{p+1}, U_{p+1} \gets \mathcal{M}_{\text{task}}(U_p, \mathcal{I}_p, \{\mathcal{J}_I\}, \{\mathcal{C}_I\})$ 
    \Comment{Generate more complex tasks based on new software knowledge and performance feedback}

\EndFor

\noindent\textcolor{gray}{\rule{\linewidth}{0.3pt}}

\State \textbf{Output:} Specialized agent policy $\pi_P$ after $P$ stages of self-evolution
\end{algorithmic}
\end{algorithm}

\section{SEAgent Self-Evolution Algorithm}

Algorithm~\ref{alg:self-evolution} presents the core self-evolution training loop of SEAgent in a specialized software environment. The procedure is divided into four major stages:

(1) \textbf{Task Initialization.} Given the initial GUI state of a target software application, the World State Model performs dense captioning to extract structural semantics (e.g., menu bar, buttons), which is used by the Curriculum Generator to create an initial set of executable tasks and an editable software guidebook.

(2) \textbf{Autonomous Exploration and Effect Evaluation.} The agent explores each task via its current policy. The World State Model then performs step-level trajectory analysis, assigning each action a feedback label—either correct ($a_T$) or incorrect ($a_F$)—and generating GUI state change captions. This produces rich supervision signals for both policy learning and downstream task generation.

(3) \textbf{Policy Update via Reinforcement Fine-Tuning.} Based on the labeled execution data, positive and negative action steps are separated. We apply Group Relative Policy Optimization (GRPO) to reinforce correct actions, and Adversarial Imitation (AI) to suppress failure-prone behaviors. The updated policy is used for the next exploration round.

(4) \textbf{Task Update.} The Curriculum Generator leverages feedback signals ($\mathcal{J}$) and GUI state transitions ($\mathcal{C}$) to propose more diverse and challenging tasks, thereby expanding the task frontier in a curriculum fashion.

This process repeats over multiple curriculum phases, ultimately yielding a specialized agent policy capable of mastering complex operations in the given software environment.

\end{document}